%% file: final_revision.tex
\documentclass{article}

\usepackage{microtype}
\usepackage{booktabs} 
\usepackage{multirow}
\usepackage{hyperref}
\usepackage{enumitem}
\usepackage{tabularx}
\usepackage{makecell}

\usepackage{wrapfig}
\usepackage[accepted]{icml2024}
\usepackage{url}

\usepackage{amsmath}
\usepackage{amssymb}
\usepackage{mathtools}
\usepackage{amsthm}

\usepackage{graphicx}
\usepackage{caption}
\usepackage{subcaption}
\usepackage[capitalize,noabbrev]{cleveref}

\theoremstyle{plain}
\newtheorem{theorem}{Theorem}[section]

\theoremstyle{definition}
\newtheorem{definition}[theorem]{Definition}

\theoremstyle{remark}

\usepackage{xcolor}

\usepackage{soul}   
\definecolor{lightgreen}{rgb}{0.8, 0.99, 0.8}  

\definecolor{lightred}{rgb}{0.99, 0.8, 0.8}    

\newcommand{\gr}[1]{{\sethlcolor{lightgreen}\hl{#1}}}
\newcommand{\re}[1]{{\sethlcolor{lightred}\hl{#1}}}
\newcolumntype{Y}{>{\raggedright\arraybackslash}X}
\usepackage[textsize=tiny]{todonotes}

\icmltitlerunning{
Token-Specific Watermarking with Enhanced Detectability and Semantic Coherence for LLMs
}

\begin{document}

\twocolumn[
\icmltitle{
Token-Specific Watermarking with Enhanced Detectability and Semantic Coherence for Large Language Models
}


\icmlsetsymbol{equal}{*}

\begin{icmlauthorlist}
\icmlauthor{Mingjia Huo}{equal,ucsd}
\icmlauthor{Sai Ashish Somayajula}{equal,ucsd}
\icmlauthor{Youwei Liang}{ucsd}
\icmlauthor{Ruisi Zhang}{ucsd}
\icmlauthor{Farinaz Koushanfar}{ucsd}
\icmlauthor{Pengtao Xie}{ucsd}
\end{icmlauthorlist}

\icmlaffiliation{ucsd}{Department of Electrical and Computer Engineering, University of California, San Diego, La Jolla, CA 92093, USA}

\icmlcorrespondingauthor{Pengtao Xie}{p1xie@ucsd.edu}

\icmlkeywords{watermarking, large language models, text generation}

\vskip 0.3in
]

\printAffiliationsAndNotice{\icmlEqualContribution} 





\begin{abstract}
Large language models generate high-quality responses with potential misinformation, underscoring the need for regulation by distinguishing AI-generated and human-written texts.
Watermarking is pivotal in this context, which involves embedding hidden markers in texts during the LLM inference phase, which is imperceptible to humans. Achieving both the detectability of inserted watermarks and the semantic quality of generated texts is challenging. While current watermarking algorithms have made promising progress in this direction, there remains significant scope for improvement. To address these challenges, we introduce a novel multi-objective optimization (MOO) approach for watermarking that utilizes lightweight networks to generate token-specific watermarking logits and splitting ratios. By leveraging MOO to optimize for both detection and semantic objective functions, our method simultaneously achieves detectability and semantic integrity. 
Experimental results show that our method outperforms current watermarking techniques in enhancing the detectability of texts generated by LLMs while maintaining their semantic coherence. Our code is available at \url{https://github.com/mignonjia/TS_watermark}.

\end{abstract}

\section{Introduction}
\label{introduction}

Large Language Models (LLMs), particularly ChatGPT, have significantly revolutionized the field of artificial intelligence (AI), bringing forth unparalleled advancements and applications with human-like capabilities~\cite{schulman2022chatgpt}. However, this rapid evolution has been accompanied by ethical challenges. Prominent among these are concerns such as their potential use in election manipulation campaigns~\cite{alvarez2023generative, wu2023large}, the creation of fake news and misleading web content~\cite{augenstein2023factuality}, and aiding academic dishonesty by facilitating cheating on homework~\cite{cheating}. In light of these issues, the detection of text generated by LLMs emerges as a critical task, underpinning the broader goals of AI ethics and safety.

Various classification-based approaches have been proposed to determine whether a text is generated by humans or LLMs~\cite{openai-detector, guo2023close, brown1992estimate}. However, as LLM-generated texts increasingly match the quality of human-generated ones, these methods are losing their robustness. For instance, OpenAI released its own AI classifier~\cite{openai-detector} in early 2023, but it was later withdrawn due to its low accuracy. 
Recently, watermarking techniques in LLMs have gained popularity~\cite{abdelnabi2021adversarial, he2022protecting, zhang2023remark, kirchenbauer23a, kuditipudi2023robust, Aaronson_wm, liu2023semantic, ren2023robust, lee2023wrote, wang2023towards, wouters2023optimizing}. These techniques embed hidden patterns in LLM-generated texts that, while imperceptible to humans, can be detected by algorithms.
\citet{kirchenbauer23a} propose an effective watermarking method, referred to as KGW in this paper. To generate a new token with a watermark, the hash of the preceding token is used as a random seed to divide the vocabulary into a green list and a red list with a fixed splitting ratio (i.e., the percentage of tokens in the green list). 
Then a constant value, known as the watermark logit, is added to the LLM-produced logits on the green list tokens. 
This adjustment increases the probability of selecting the `green' tokens as the new token.  
To identify the presence of a watermark, a statistical test is carried out. This involves counting the number of green list tokens in the generated text. A higher incidence of green tokens suggests that the text is likely to contain a watermark.

\begin{figure*}
    \centering
    \includegraphics[trim={0cm 2.95cm 2.4cm 4.3cm}, clip,  width=0.8\textwidth]{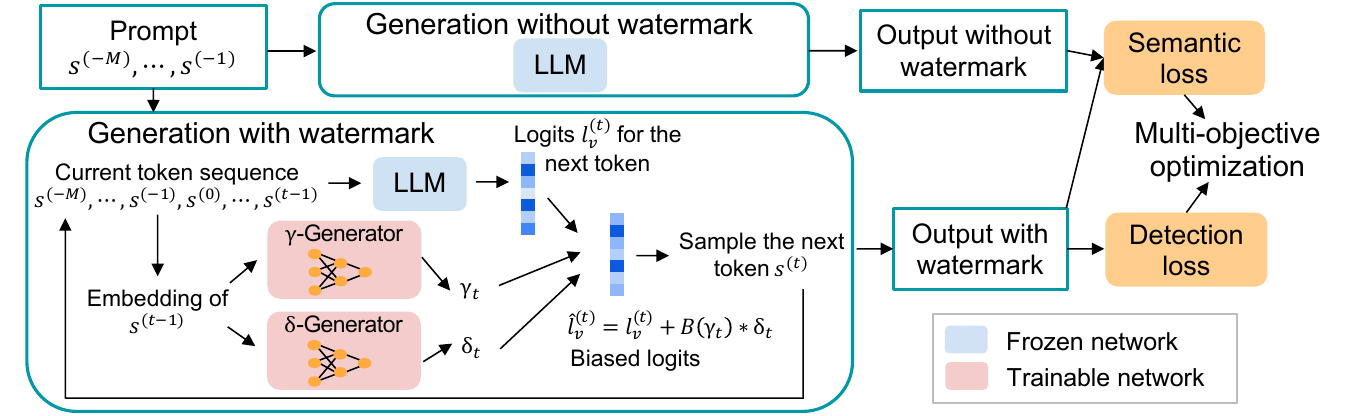}
    \caption{The training procedure is as follows: During the LLM text generation, we utilize the \(\gamma\)-generator and \(\delta\)-generator to modify the probability of each token before sampling the next one. The parameters of these networks are learned through optimization of the detection loss (Eq.~\ref{eq:detection-loss}) and semantic loss (Eq.~\ref{eq:semantic-loss}) within a multi-objective optimization framework.
    }
    \label{fig:workflow}
\end{figure*}

The design of KGW emphasizes easy detection of watermarked texts. However, this approach often compromises the semantic coherence of the texts, as highlighted in \citet{lee2023wrote}.
One primary cause of this issue is that KGW uses a constant splitting ratio and watermark logit across all tokens, without taking into account the context and semantics of the specific token being generated. 
For instance, given the prefix  ``The sun rises in the",  it is highly probable that the next token to be generated should be ``east''. But KGW's watermarking mechanism might not choose ``east'' if the splitting ratio is low and the watermark logit is high. 
A lower splitting ratio results in a smaller number of words being added to the green list, thus significantly reducing the chances of including the word ``east'' in it. On the other hand, a higher watermark logit raises the probability of choosing a token from this list. Consequently, it becomes highly unlikely for ``east'' to be selected as the next token for generation, which will significantly impact the semantic coherence of the text. 
To mitigate this issue, adjusting the splitting ratio and watermark logit is necessary, perhaps by increasing the splitting ratio or decreasing the watermark logit. Unfortunately, KGW lacks such an adaptive mechanism.  Conversely, if KGW  utilizes a low, constant splitting ratio and  a small, uniform watermark logit for every token to maintain semantic integrity, this approach may hinder the watermark's detectability.
Hence, it is crucial to adaptively assign token-specific values for these hyperparameters, simultaneously ensuring high detectability while maintaining semantic integrity. 
Some works have been proposed to improve the semantics of watermarked texts. 
\citet{lee2023wrote} proposes a selective watermarking strategy to preserve semantics by adding watermarks only to high-entropy tokens while preserving the original logits for low-entropy tokens.
\citet{kuditipudi2023robust} reduce the semantics distortion by ensuring an unbiased distribution of tokens before and after watermarking, using an exponential minimum sampling strategy which can positively influence semantics. 
However, these works face challenges in enhancing both detectability and semantic coherence at the same time: improving one aspect frequently compromises the other.

To address the limitations of current methods, we introduce a novel method (Figure~\ref{fig:workflow}) that simultaneously achieves two primary goals: preserving the semantic integrity of generated texts and ensuring the effectiveness of watermark detection. Our method accomplishes the goals by dynamically adjusting the splitting ratio and watermark logit, controlled by two trainable 
light-weight networks, for each token during its generation. 
These networks process the representation of the previous token to determine the optimal splitting ratio and the appropriate watermark logit for the next token, supervised by two loss functions: 
1) Watermark detectability via a one-sided z-test~\cite{kirchenbauer23a}, which quantifies the presence of green tokens in the generated text. Since this metric is inherently non-differentiable, we introduce a differentiable surrogate that allows for direct optimization through gradient-based techniques during training. 
2) Semantic coherence of watermarked texts, for which we measure the cosine similarity between SimCSE~\cite{gao2021simcse} embeddings of watermarked and non-watermarked texts.
We develop a multi-objective optimization framework that aims to achieve both objectives concurrently.
Our method is geared towards identifying Pareto optimal solutions, where improving one objective does not detrimentally affect the other. This balanced approach ensures the effectiveness of watermarking while maintaining the semantic quality of the generated texts.

Our main contributions include:
\begin{itemize}
    \item 
    We introduce a novel watermarking method for LLMs that improves both detectability and semantic coherence in generated texts. Unlike earlier methods, which often face challenges in achieving these objectives simultaneously, our approach employs multi-objective optimization to achieve both of them concurrently.
    
    \item Our method employs two lightweight networks to dynamically determine token-specific splitting ratios and watermark logits, avoiding uniform values across all tokens. 
    
    \item Our method has undergone comprehensive evaluation, showing superior performance in both detectability and semantic quality compared to leading baselines. 
\end{itemize}

\section{Related Works}

\subsection{Watermarking Methods}

Watermarking methods for large language models can be divided into three categories: rule-based watermarking, neural watermarking and inference-time watermarking. Rule-based watermarking embeds watermarks using text transformations while ensuring that the overall semantic coherence is not disturbed. These transformations involve altering lexical properties~\cite{he2022protecting}, manipulating linguistic features~\cite{he2022cater, yoo2023robust}, or substituting synonyms~\cite{munyer2023deeptextmark, yang2023watermarking}. A significant limitation of rule-based methods is that they are vulnerable to attacks (e.g., replacing words with synonyms).  
Neural watermarking employs neural networks to embed watermarks into texts and decode them. \citet{abdelnabi2021adversarial} leverage a Transformer~\cite{vaswani2017attention} encoder to embed a binary watermark message, which is then extracted by a Transformer decoder. 
A Transformer-based discriminator is employed to preserve the original semantics of the text while applying watermarking. REMARK-LLM~\cite{zhang2023remark} employs a message encoding network to embed an LLM-generated text and its signature into a watermarked version. It then utilizes a message decoder to extract the LLM signature from this watermarked text. 
Neural watermarking methods often involve complicated neural networks to insert watermarks, incurring high computational costs during text generation and watermark detection.   

Inference-time watermarking methods  insert statistical signals into  model logits during inference to improve detectability. KGW~\cite{kirchenbauer23a} comes under this umbrella. However,  text  watermarked by this method suffers from reduced semantic coherence. A series of works have been proposed to address this issue. 
\citet{wang2023towards} 
introduce a constraint stipulating that the perplexity of watermarked text must not exceed the perplexity of the original text, as produced by the same language model without watermarking, by more than a small margin.
However, this strategy is developed through a series of approximations to ensure practical applicability. These approximations can lead to semantic disparities between the two texts. 
\citet{chen2023xmark} propose to split  the vocabulary into a green list and red list in a semantically more balanced manner so that any token in the red list can be replaced by a token in the green list. 
However, their splitting method is based on LLAMA2~\cite{touvron2023llama}, which is  computationally expensive. 
Furthermore, \citet{liu2023semantic}  introduce a semantically invariant watermarking approach, aimed to improve attack and security robustness. They employ Compositional-BERT~\cite{chanchani2023composition} to extract semantic representations of  preceding tokens and produce watermark logits from these representations. However, this method incurs high computation overhead during inference due to the utilization of Compositional-BERT. \citet{wouters2023optimizing} present a closed-form solution for optimizing  watermarking  logits  to enhance semantic integrity, with a predetermined splitting ratio. However, this solution relies on assumptions that may not hold in real-world scenarios, such as all tokens following identical distribution, and the absence of distribution shifts post-watermarking. Additionally, their analysis does not extend to determining an optimal splitting ratio - a factor that is critical for balancing detectability and semantic preservation. To enhance semantic coherence of watermarked computer programs,  SWEET~\cite{lee2023wrote} proposes to 
selectively insert watermarks into high-entropy tokens instead of every token. While this method is effective for watermarking code, it demonstrates weak detection capabilities in texts across general domains.

\subsection{Multi-Objective Optimization}

Multi-objective optimization (MOO)~\cite{deb2016multi} addresses the challenge of simultaneously optimizing several objectives which often conflict with each other. The Multiple-gradient Descent Algorithm (MGDA)~\cite{desideri2012multiple} is a notable gradient-based approach designed for solving MOO problems. MGDA aims to identify a single gradient direction theoretically capable of optimizing all objectives concurrently. 
If the gradients are not normalized, the direction of the optimization is expected to be mostly influenced by the gradients of small norms~\cite{desideri2012multiple}. 
To mitigate this, various normalization methods have been developed~\cite{chen2018gradnorm, milojkovic2019multi}, striving for a more equitable consideration of all involved gradients.  In our work, we employ MGDA to identify the Pareto optimal solutions between detectability and semantic coherence.

\section{Preliminaries}
\label{preliminaries} 
Our work is built upon an inference-time watermarking strategy introduced in~\citet{kirchenbauer23a}. 
This watermarking technique is a two-stage process: first, embedding a watermark into the text during its generation, and second, identifying the presence of this watermark in the text. 
During the generation of token $s^{(t)}$, the hash of the preceding token $s^{(t-1)}$ serves as a random seed. This seed is used to divide the vocabulary $\mathcal{V}$ into a green list, which contains a fraction $\gamma$ of the vocabulary, and a red list, containing the remaining $(1 - \gamma)$ fraction of the vocabulary. The parameter $\gamma$, known as the splitting ratio, determines the size of the green list relative to the total vocabulary. Next, a constant watermark logit, denoted as $\delta$, is added as a bias to the logits of green list tokens. The sampling of the next token is then based on these adjusted logits, softly prompting the use of green list tokens. 

The process of detecting watermarked text does not require access to the LLM  that originally generated it. Knowing the specific hash function and random number generator utilized is sufficient to reproduce the red and green lists associated with each token. 
The detection process involves testing the null hypothesis \( H_0 \)   that the text was generated without knowledge of the green-red list rule. This is assessed using a one-sided z-test, with a z-score  \( z = (|s|_G - \gamma T) / \sqrt{T\gamma(1-\gamma)} \) under \( H_0 \), where \( |s|_G \) represents the count of green list tokens in the watermarked text and $T$ denotes the length of the text. A watermark is considered successfully detected if this test leads to the rejection of the null hypothesis, which occurs when the calculated z-score exceeds a predetermined threshold. 

\section{Method}
\subsection{Overview}
We present a novel watermarking method for LLMs, designed to optimize two key aspects: detectability and semantic coherence. Detectability is assessed using a one-sided z-test that calculates a z-score based on the count of  green tokens in the text. Semantic coherence is evaluated by measuring the cosine similarity between the embeddings of watermarked and non-watermarked texts.  These measures are  controlled by two hyperparameters: the split ratio $\gamma$ and watermark logit $\delta$. Their values vary for different tokens to reflect tokens' unique characteristics.

To specify token-specific values of $\gamma$ and $\delta$, we employ two light-weight networks: a $\gamma$-generator $G_\gamma$ and a $\delta$-generator $G_\delta$. The $\gamma$-generator processes the representation of a previous token $t-1$ to determine the split ratio $\gamma_{t}$ for  token $t$, and similarly, the $\delta$-generator operates for the watermark logits. These networks are optimized through a specialized multi-objective optimization framework~\cite{desideri2012multiple, NeurIPS2018_Sener_Koltun}, aiming to simultaneously enhance detectability and semantic coherence. 
Figure~\ref{fig:workflow} shows an overview of our proposed method.

\subsection{Network Design}\label{subsec:networks}

In this section, we provide a detailed description of  the $\gamma$-generator  and $\delta$-generator. 

\paragraph{$\gamma$-Generator.}\label{gamma-generator}

The $\gamma$-generator is a lightweight multi-layer perceptron.  It takes the embedding of the preceding token $s^{(t-1)}$ as input and generates a splitting ratio $\gamma_t\in(0,1)$ for token $t$. 
Then we  define a Bernoulli distribution parameterized by $\gamma_t$, denoted as $B(\gamma_t)$, to split the vocabulary $\mathcal{V}$ into a token-specific green list and  red list. Specifically, for each token $v$ in $\mathcal{V}$, we draw a sample from this distribution, $y_v^{(t)} \sim B(\gamma_t)$, to determine whether the token belongs to the green list.  
The token $v$ is assigned to the green list if the sampled value $y_v^{(t)}$ is 1, and to the red list if it is 0.

However, the sampling process from a Bernoulli distribution is non-differentiable, which prevents the gradient-based updating of the parameters in $G_\gamma$. To address this issue, we utilize the Gumbel-Softmax method for differentiable sampling~\cite{jang2017categorical}.
Specifically, for each token \( v \) in \( \mathcal{V} \), we estimate the probability that it belongs to the green list, denoted as \( \hat{y}_{v}^{(t)} \), using the following formula:
\vspace{-2ex}

\small
\begin{align}
\hat{y}_{v}^{(t)} = \frac{\exp\left(\frac{\log(\gamma_t) + g_{0}}{\tau}\right)}{\exp\left(\frac{\log(\gamma_t) + g_{0}}{\tau}\right) + \exp\left(\frac{\log(1-\gamma_t) + g_{1}}{\tau}\right)}.
\end{align}
\normalsize

Here, $g_{0}$ and $g_{1}$ are i.i.d samples from Gumbel$(0,1)$\footnote{The Gumbel$(0, 1)$ distribution can be sampled using inverse transform sampling by drawing $u\sim$ Uniform$(0, 1)$ and computing $g=-\log(-\log (u))$.}.
Here $\tau$ serves as a temperature parameter. As $\tau$ approaches 0, the distribution becomes increasingly sharp, which results in $\hat{y}_{v}^{(t)}$ more closely approximating ${y}_{v}^{(t)}$. We utilize $\hat{y}_{v}^{(t)}$, a differentiable approximation of ${y}_{v}^{(t)}$, to softly split the vocabulary into red and green token lists.

\paragraph{$\delta$-Generator.}
\label{delta-generator}

Similarly to the $\gamma$-generator, the $\delta$-generator is also a lightweight multi-layer perceptron, which takes the embedding of the preceding token $s^{(t-1)}$ as input and generates a watermark logit $\delta_t\in \mathbb{R}^+$ for token $t$ to bias the green list tokens.   
Recall from Sec.~\ref{gamma-generator}, $\hat{y}_v^{(t)}$ is the probability that a token, $v \in \mathcal{V}$, belongs to the green list. The watermark logit $\delta_t$ is used to bias the model logit  $l_v^{(t)}$ for token $v$ as follows:  
\begin{equation}
    \hat l_v^{(t)} = l_v^{(t)} + \hat{y}_v^{(t)} \cdot \delta_t
    \label{eq:modified_logits}. 
\end{equation}

This formulation modifies the logit of token $v$ by adding an appropriate amount of \(\delta_t\), based on the likelihood of the token being in the green list. These adjusted logits, \(\{\hat l_v^{(t)} | v\in \mathcal{V}\}\), are transformed into a probability vector using SoftMax, which is then used to sample the next token.

\subsection{Training Objectives}
\label{subsec:optim}

Our goals are to ensure both strong watermark detection ability and high semantic coherence after adding the watermark. To achieve these goals, we leverage two training objectives: a detection loss measured by a differentiable approximation of $z$-score and a semantic loss measured by the cosine similarity of watermarked and unwatermarked sentence embeddings.

\paragraph{Detection Loss.}

As described in Sec.~\ref{preliminaries}, KGW ~\cite{kirchenbauer23a} uses a constant green list ratio $\gamma$, and applies a one-sided z-test for watermark detection,
represented as \((|s|_G - \gamma T) / \sqrt{T\gamma(1-\gamma)} \). 
However, our method introduces a novel variation: $\gamma$ is dynamically adapted for each token based on the semantics of the preceding token. 
We need to modify the z-score calculation to account for this dynamic adaptation of $\gamma$. Thus, the mean and the standard deviation of the distribution under $H_0$ need to be estimated. The generated $t^{\text{th}}$ token can be either a red list token or a green list token. Under hypothesis $H_0$, this categorization follows a Bernoulli distribution  $B(\gamma_t)$ parameterized by $\gamma_t$. To elucidate this, we define a random variable $X_t \sim B(\gamma_t)$, where $X_t = 1$ signifies a green list token and $X_t = 0$ a red list token.
The mean of $X_t$ is thus $\gamma_t$. Consequently, the total count of green list tokens in the entire sequence of length $T$ is represented by $\sum_{t=0}^{T-1}X_t$.
\vspace{1ex}

\begin{theorem}
Consider $T$ independent Bernoulli random variables $X_1,\ldots,X_T$, each with means $\mu_1,\ldots, \mu_T$, \( 0 < \mu_t < 1 \) $\forall t \in 1, \ldots, T$. The sum of these variables, $\sum_{t=1}^T X_t$, follows a Poisson binomial distribution. When $T$ is sufficiently large, this distribution can be approximated by a Gaussian distribution with mean: $\sum_{t=1}^T \mu_t$ and variance: $\sum_{t=1}^T \mu_t(1-\mu_t)$. 
\label{theorem:1}
\end{theorem}

Using Theorem~\ref{theorem:1}, when $T$ is sufficiently large, the z-score under the null hypothesis, $H_0$, can be approximated by the following expression:
\begin{equation}
    z = \frac{{|s|}_G - \sum_{t=1}^T \gamma_t}{\sqrt{\sum_{t=1}^T \gamma_t(1-\gamma_t)}}. 
\label{eq:z-score}
\end{equation}

Our goal is to enhance detectability by optimizing this  z-score. However, in its current formulation, the z-score is not differentiable with respect to the parameters of the $\gamma$- and $\delta$-generator, due to the term ${|s|}_G$. Therefore, we propose a relaxed formulation, by relaxing  ${|s|}_G$, the number of green list tokens, as  $\sum_{t=1}^T p_{gr}^{(t)}$. Given a red-green token list during the generation of the $t^{\text{th}}$ token, $p_{gr}^{(t)}$ represents the probability of selecting a green token, as determined by the modified logits in Eq.~\ref{eq:modified_logits}. This probability is computed as the summation of the probabilities of all tokens in the green list, where these individual probabilities are  calculated using the Softmax function applied to the modified logits. The relaxed z-score is: 
\begin{equation}
    \hat{z} = \frac{\sum_{t=1}^T p_{gr}^{(t)} - \sum_{t=1}^T \gamma_t}{\sqrt{\sum_{t=1}^T \gamma_t(1-\gamma_t)}}. 
\label{eq:detection-loss}
\end{equation}

It is differentiable with respect to the parameters of the $\gamma$ and $\delta$ generators. The objective of our approach is to maximize $\hat{z}$, thereby increasing detectability. Consequently, the detectability loss $L_D$ is defined as $L_D = -\hat{z}$, which we aim to minimize.
 
\paragraph{Semantic Loss.}
\label{semantic loss}

To maintain the semantic integrity of watermarked texts, we aim for a high degree of semantic resemblance to their original, non-watermarked counterparts generated by LLMs. To evaluate the semantic similarity between the two versions of the text, we compute the cosine similarity of their latent embeddings, which are  generated by the RoBERTa-base model~\cite{liu2019roberta} from SimCSE~\cite{gao2021simcse}, denoted by $f_\theta$. The SimCSE approach pretrains the RoBERTa-base model to generate sentence embeddings under a contrastive learning framework~\cite{hadsell2006dimensionality} so that the cosine similarity of the embeddings can indicate their semantic similarity. 
Specifically, considering $s$ and $s_w$ as the non-watermarked and watermarked LLM-generated sentences, respectively, we improve the semantic integrity of $s_w$ by minimizing the following semantic loss: 
\begin{equation}
    L_S = - \cos_{\text{sim}}(f_\theta(s), f_\theta(s_w)), 
\label{eq:semantic-loss}
\end{equation}
where \( \cos_{\text{sim}}(\cdot , \cdot) \) represents the cosine similarity operation. The optimization variables are the weight parameters in the $\gamma$- and $\delta$-generator networks since the watermarking of $s_w$ is controlled by these networks. Since the RoBERTa-base model $f_\theta$ and the LLM that generates $s_w$ share the same tokenizer, we directly use the embedding generated by the LLM as the input to $f_\theta$, making the operation differential w.r.t.~the $\gamma$- and $\delta$-generator networks.

\subsection{Multi-Objective Optimization}\label{sec:moo}
As explained earlier, the detection loss $L_D$ and semantic loss $L_S$ have a competing relationship: solely decreasing one of them often leads to the increase of the other. To ensure concurrent reduction of both losses, we formulate a  multi-objective optimization problem: 
\begin{equation}
    \min_{G_\gamma, G_\delta} L_D(G_\gamma, G_\delta)\text{ and }\min_{G_\gamma, G_\delta} L_S(G_\gamma, G_\delta).
\label{eq:moo}
\end{equation}
The goal of this formulation is to achieve Pareto optimality as defined below.

\vspace{1ex}
\begin{definition}{(Pareto Optimality)}
    \begin{enumerate}[label=(\alph*).]
        \item A solution ($G_\gamma, G_\delta$) dominates a solution ($\overline{G_\gamma}, \overline{G_\delta}$) if 
        \begin{itemize}
            \item $L_D(G_\gamma, G_\delta) \le L_D(\overline{G_\gamma}, \overline{G_\delta})$
            \item $L_S(G_\gamma, G_\delta) \le L_S(\overline{G_\gamma}, \overline{G_\delta})$
        \end{itemize}
        and at least one inequality is strict.
        \item A solution $(G_\gamma, G_\delta)$ is called Pareto optimal if there exists no other solution that dominates it.
    \end{enumerate}
\end{definition} 
We employ the multiple-gradient descent algorithm  (MGDA)~\cite{desideri2012multiple, NeurIPS2018_Sener_Koltun}, to solve the multi-objective optimization problem described in Eq.(\ref{eq:moo}). Please refer to Appendix~\ref{sec:appdix_moo} for details.

\subsection{Watermark Generation and Detection}\label{subsec:inf}

In this section, we introduce  how to generate and detect a watermark using the trained $G_\gamma$ and $G_\delta$. The text generation process at inference time is similar to the procedure used in KGW. The primary difference is that we utilize a dynamic $\gamma_t$ and $\delta_t$ for each token, outputted by the trained $G_\gamma$ and $G_\delta$, respectively, with the inputs being the embedding of the preceding token.
Detection is conducted through one-sided z-test, specifically by using the z-score as defined in Eq.(\ref{eq:z-score}).
Given a generated text, $G_\gamma$ is utilized to compute $\gamma_t$ necessary for calculating the z-score. Compared with the original KGW method, the only additional requirement during watermark detection is the embedding matrix of the LLM, which is used to compute the embedding of the preceding token. 

\section{Experiments}
In our experiments, we insert watermarks to two prominent LLMs: OPT-1.3B~\cite{zhang2022opt}, LLAMA2-7B, \textcolor{black}{13B and 70B}~\cite{touvron2023llama}. The evaluation of watermarking is based on three aspects: the trade-off between detectability and semantic integrity, computational complexity, and robustness against attacks of the watermarks. We also perform an analysis of our learned \(\delta\) and \(\gamma\), and present an ablation study in Appendix~\ref{sec:appdix-weighted}. 
Training details including the hyperparameter settings can be found in Appendix~\ref{sec:appdix_exp}.

\subsection{Experimental Settings}\label{sec:exp_setting}

\paragraph{Dataset and Prompt.}

Following~\citet{kirchenbauer23a}, we utilize texts from the news-like subset of the C4 dataset~\cite{2019t5} to insert watermarks. For each text from this dataset, the last 200 tokens are truncated and designated as the `baseline completion' (i.e., human-written texts). The remaining tokens from the start of the text are considered the `prompt'. Conditioned on this prompt,  the LLM generates a token sequence equivalent in length to the baseline completion, incorporating watermarks within this generation.  
Following~\citet{kirchenbauer23a}, we filter the dataset to include texts ranging from 500 to 1000 tokens. This dataset is then divided randomly into three subsets: 6,400 texts for training, 500 for validation, and another 500 for testing.

\paragraph{Evaluation Metrics.}

The objective of a watermarking algorithm is to effectively identify texts generated by LLMs, i.e., achieving a high true positive rate (TPR),  while not classifying texts created by humans as LLM-generated, i.e., with a low false positive rate (FPR).  By raising the threshold for the z-score in detection processes, we can effectively reduce the FPR. 
We set this threshold to maintain FPRs at two different levels: 0\% and 1\%. This is achieved by identifying the threshold corresponding to the top 0\% (or 1\%) of the z-scores of all `baseline completions' on the test set. Using these thresholds, we then evaluate the TPR for watermarked texts generated by LLMs.
%
Additionally, we evaluate the semantic coherence between texts generated with and without watermarks by LLMs. We measure this by calculating the cosine similarity of their sentence embeddings, obtained using the SimCSE model (as detailed in Sec.~\ref{semantic loss}). The metric assesses the impact of watermarking on the quality of generated text. For each FPR level, we plot a trade-off curve between TPR and semantic similarity to assess the performance of the watermarking algorithms. 

We compare our methods against KGW~\cite{kirchenbauer23a}, SWEET~\cite{lee2023wrote}, MultiBit~\cite{wang2023towards}, SIR~\cite{liu2023semantic}, and EXP-edit~\cite{kuditipudi2023robust}. Both SWEET and SIR rely on prompts during detection, which is impractical in many real-world scenarios. To ensure a fair comparison, we also include variants of these baselines that do not use prompts during detection, namely $\text{SWEET}_\text{NoPrompt}$ and $\text{SIR}_\text{NoPrompt}$.
We plot trade-off curves of TPR and semantic similarity for the two settings, FPR=0\% and FPR=1\%. 
We vary \(\delta\) within the appropriate range for each baseline to generate different pairs of TPR and semantic similarity. For KGW, SWEET, and $\text{SWEET}_{\text{NoPrompt}}$, we set \(\gamma = 0.25\). 
For MultiBit, we use message lengths of 9 and 7 for which the FPR is 0\% and 1\%, respectively. Multinomial sampling with a temperature of $1.0$ and Top-$k$ of 50 is used for text generation.
For our method, we generate different pairs of TPR and semantic similarity using different initializations of the $\gamma$- and $\delta$-generator networks. Then we fit a curve to these pairs (i.e., points) for each method and plot the curves to form the Pareto frontier~\cite{giagkiozis2014pareto}, which represents the best trade-off curve of TPR and semantic similarity. For more details, please refer to Appendix~\ref{sec:appdix_exp}.

\subsection{Results on Detectability and Semantic Coherence}

\begin{figure}[t]
    \centering
    \includegraphics[width=0.45\textwidth]{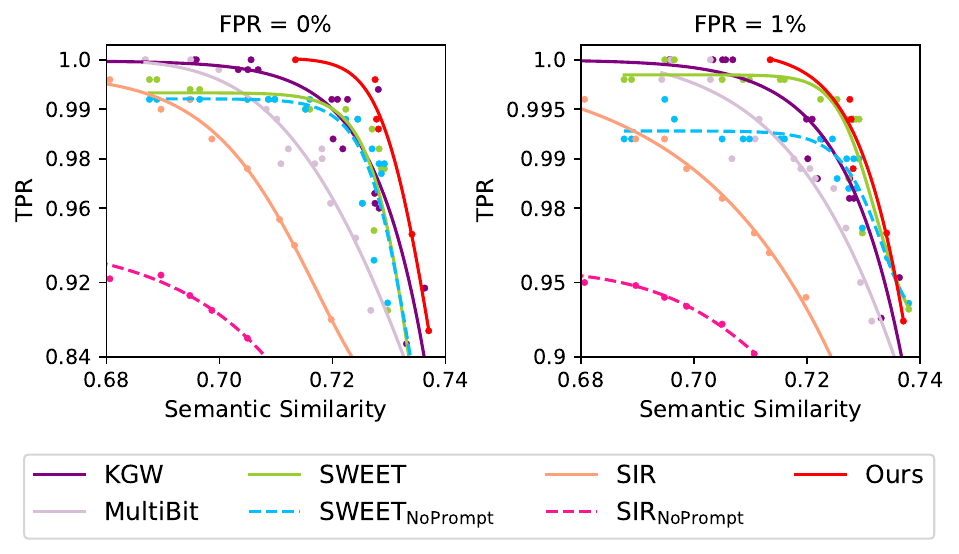}
    \caption{Comparison of the trade-off for semantic integrity and detectability of different methods applied to OPT-1.3B.}
    \label{fig:main_result}
\end{figure}

\begin{figure}[t]
    \centering
    \begin{subfigure}[t]{0.45\textwidth}
        \includegraphics[width=\textwidth]{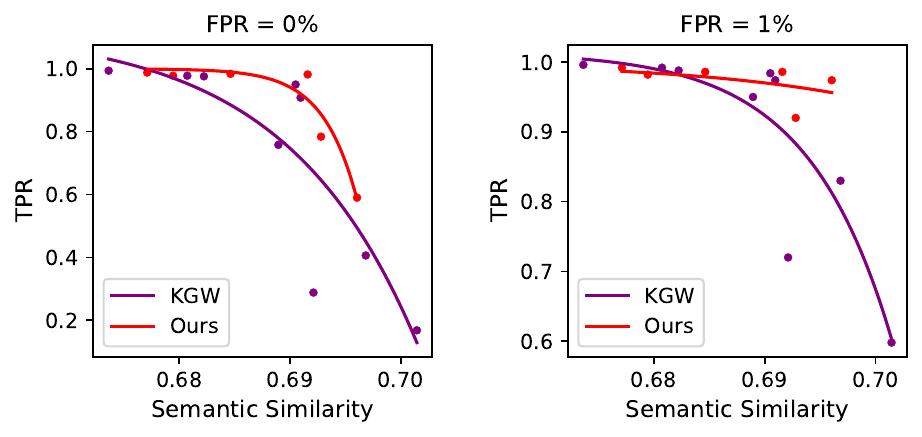}
        \caption{LLAMA2-7B}
    \end{subfigure}
    \begin{subfigure}[t]{0.45\textwidth}
        \includegraphics[width=\textwidth]{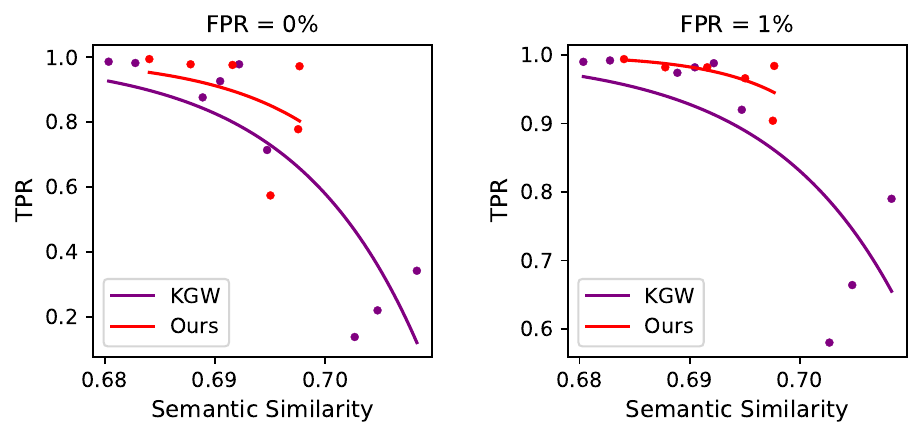}
        \caption{LLAMA2-13B}
    \end{subfigure}
    \begin{subfigure}[t]{0.45\textwidth}
        \includegraphics[width=\textwidth]{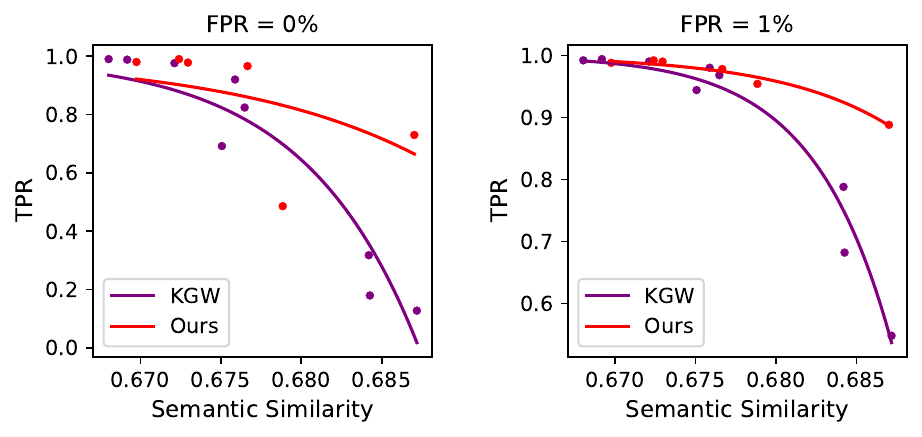}
        \caption{LLAMA2-70B}
    \end{subfigure}
    \caption{Performance of Ours (trained on OPT-1.3B) and KGW when applied to LLAMA2 7B, 13B, and 70B.}
    \label{fig:llama}
\end{figure}

Figure~\ref{fig:main_result} and~\ref{fig:llama} show results on OPT-1.3B and LLAMA2 (7B,  13B and 70B), respectively, which highlight several key observations when comparing with the baselines. 

\textbf{Comparison with KGW~\cite{kirchenbauer23a}}: Our approach improves the Pareto frontier compared to the KGW baseline, attributable to two key factors. The first factor is that  our method uniquely learns token-specific splitting ratios and watermark logits, which take into account the distinct context and semantics of each token. This is critical because the number of semantically appropriate tokens that can be chosen as the next token in the sequence can vary substantially at different time steps during text generation, depending on the context.  In scenarios where this number is small (such as the example in Sec.~\ref{introduction}), it is essential to lower the splitting ratio and watermark logit to reduce the watermark strength. This adjustment increases the likelihood of selecting these valid choices as the next token, thereby maintaining semantics. 
Conversely, in situations where this number is large, increasing the splitting ratio and watermark logit to an \emph{appropriate} level can improve detectability while posing minimal risk to semantic coherence. 
Our method has the adaptability that allows for token-specific adjustments in splitting ratios and watermark logits, while KGW employs uniform values across all tokens.  The second factor distinguishing our approach is the incorporation of multi-objective optimization, which enables simultaneous maximization of detectability and semantic integrity. This is achieved by concurrently optimizing a differentiable detection loss and a semantic loss. In contrast, the KGW method cannot explicitly optimize for these two objectives together.

\textbf{Comparison with SWEET~\cite{lee2023wrote}}: Our method achieves a better Pareto frontier than SWEET at 0\% FPR while maintaining similar performance at 1\% FPR. At 0\% FPR, SWEET is notably less effective compared to our approach and KGW, and does not achieve 100\% TPR. This may be due to its selective watermarking strategy, which targets only high-entropy words and leaves low-entropy words un-watermarked (\(\delta = 0\)). For instance, at \((\gamma, \delta) = (0.25, 3.0)\), an analysis of LLM-generated texts that SWEET failed to detect at 0\% FPR shows that, on average, only 7 out of 200 tokens are high-entropy and eligible for SWEET watermarking. This limited number of watermarkable tokens reduces SWEET's detectability even with high \(\delta\) (see Appendix~\ref{sec:sweet-dis}). In contrast, both our method and KGW achieve a 100\% TPR in high \(\delta\) regions. Furthermore, $\text{SWEET}_{\text{NoPrompt}}$ underperforms SWEET, indicating the method's dependence on prompts, which is impractical.

\textbf{Comparison with MultiBit~\cite{wang2023towards}}: Our method achieves a superior Pareto frontier compared to MultiBit. MultiBit embeds multi-bit information into LLM-generated texts, detecting a watermark if the decoded message matches the embedded one. However, embedding multi-bit information often reduces text quality. To mitigate this, Balance-Marking strategy is introduced to decrease the perplexity of watermarked texts. However, this method is developed through a series of approximations to ensure practical applicability, which might limit its effectiveness.
In contrast, our method directly maximizes differentiable metrics of semantic coherence and detectability through multi-objective optimization, inherently improving them.

\textbf{Comparison with SIR~\cite{liu2023semantic}}: Our method achieves an improved Pareto frontier than SIR. SIR primarily aims to enhance attack and security robustness, lacking a direct approach to boost text quality. Conversely, our method employs multi-objective optimization to effectively enhance both text quality and watermark detectability simultaneously. Additionally, $\text{SIR}_{\text{NoPrompt}}$ significantly underperforms compared to SIR, indicating a strong dependence on prompts while detection. SIR is less robust than SWEET in the no-prompt scenario, as it exhibits a greater performance degradation without prompts compared to SWEET.

\textbf{Comparison with EXP-edit~\cite{kuditipudi2023robust}}: EXP-edit's generation process, which employs exponential minimum sampling, is pseudo-random and becomes deterministic with a watermark key, facilitating detection.
Unlike ours and other KGW-based methods, which shift the distribution using \(\delta\), EXP-edit does not alter the output distribution of LLMs, making it indistinguishable. Since EXP-edit does not have a \(\delta\) parameter to vary and plot the Pareto frontier, we present the results in Table~\ref{table:exp-edit}. 
Our method outperforms EXP-edit in both detectability and SimCSE, demonstrating that learning token-specific parameters to watermark enables appropriately shifting the output distribution to enhance detectability without significantly affecting semantics. This offers more freedom to effectively embed watermark compared to EXP-edit, which lacks this capability~\cite{piet2023mark}.

\begin{table}[t]
\centering
\caption{\textcolor{black}{Comparison of EXP-edit and Our Method}}
\small
\resizebox{\columnwidth}{!}{%
\begin{tabular}{l c c c}
\toprule
\textbf{Method} & \textbf{TPR @ 0\%} & \textbf{TPR @ 1\%} & \textbf{SimCSE} \\ 
\midrule
EXP-edit & 0.922 & 0.996 & 0.655 \\ 
EXP-edit (Top-$k$=50) & 0.968 & 0.996 & 0.677 \\ 
Ours (Top-$k$=50) & \textbf{1.000} & \textbf{1.000} & \textbf{0.713} \\ 
\bottomrule
\end{tabular}
}
\label{table:exp-edit}
\end{table}

Since our decoding strategy uses the default Top-$k$ value of 50, we modified the original EXP-edit generation process, which did not implement Top-$k$, to sample from the top 50 tokens, naming it EXP-edit (Top-$k$=50). We observe that our method still outperforms EXP-edit (Top-$k$=50) in terms of detectability and SimCSE. Moreover, indistinguishable methods like EXP-edit, which rely on pseudo-random sampling during generation, cannot easily extend to other decoding methods like beam search, which do not involve randomness. In contrast, KGW-based methods can be applied on top of any decoding method such as beam search~\cite{kirchenbauer23a}. See Appendix~\ref{sec:exp-editvsours} for further discussion.

\textbf{Computation Overhead}: We evaluate the computational time of our method and the baselines for text generation and detection in Table~\ref{tab:speed}. Our method achieves higher speeds than EXP-edit, SIR, and MultiBit, while achieving speeds comparable to KGW, SWEET, and No Watermarking.

\begin{table}[t]
\centering
\caption{Generation and detection speed on OPT-1.3B for generating 200 tokens, measured in seconds. We also show memory utilization in Table~\ref{tab:memory}.}
\small
\begin{tabular}{l|c|c}
\toprule
\textbf{Method} & \textbf{Generation (s)} & \textbf{Detection (s)} \\
\midrule
No Watermark & 3.220 & - \\
KGW & 3.827 & 0.067 \\
SWEET & 4.030 & 0.127 \\
EXP-edit & 24.693 & 155.045 \\
SIR & 8.420 & 0.337 \\
MultiBit & 6.500 & 0.610 \\
Ours & 3.946 & 0.166 \\
\bottomrule
\end{tabular}
\label{tab:speed}
\end{table}

\textcolor{black}{\textbf{Generalizability}:}
Our method demonstrates good generalizability across LLMs of different sizes. As shown in Figure~\ref{fig:llama}, our model ($\gamma$- and $\delta$-generator networks), initially trained on OPT-1.3B, demonstrates a better Pareto frontier \textcolor{black}{when applied to LLAMA2 7B, 13B and 70B.} This adaptability is likely because our method learns the watermarking parameters that reflect the general nature of language itself (see Sec.~\ref{analysis-of-delta-gamma} and Appendix~\ref{sec:example} for discussions), not just on the specific details of the LLMs.

\subsection{Analysis of Learned $\delta$ and $\gamma$}
\label{analysis-of-delta-gamma}

\begin{figure}[t]
    \centering
    \includegraphics[width=\linewidth]{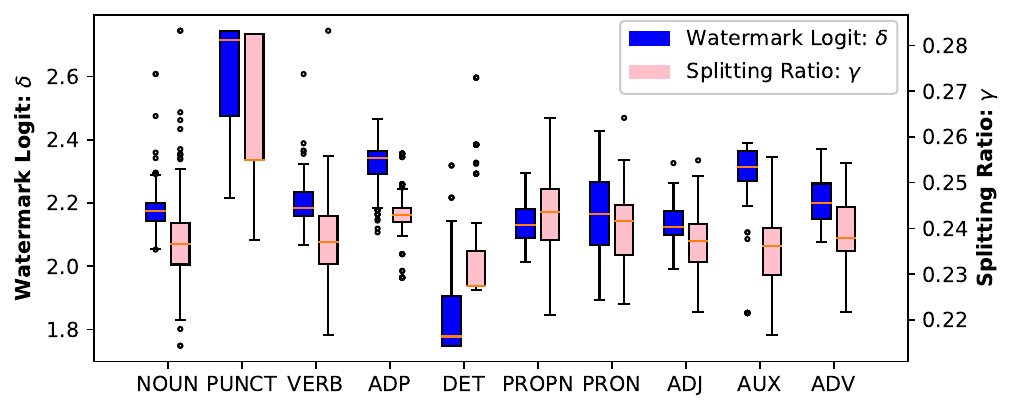}
    \caption{Distribution of watermark logit \(\delta\) (left y-axis) and splitting ratio \(\gamma\) (right y-axis) across different part-of-speech categories of the \emph{preceding} token.}
    \label{fig:delta_gamma}
\end{figure}

In this section, we examine the values of \(\gamma\) and \(\delta\) learned by our method  for different tokens. For each part-of-speech (POS) category, we calculate the mean and standard deviation of  \(\gamma\) and \(\delta\) values generated based on  \textit{preceding} tokens that are tagged with this category. This analysis is conducted on  ten texts, each containing 200 tokens that have been watermarked.  The results are presented in Figure~\ref{fig:delta_gamma}.

\textcolor{black}{One observation is that when the preceding token is an adjective (ADJ) or a determiner (DET), \(\gamma\) and \(\delta\) tend to be assigned lower values. 
This pattern is notable because ADJs and DETs typically precede nouns, as detailed in Appendix~\ref{sec:example}. 
Reducing \(\gamma\) leads to the selection of fewer green tokens, while simultaneously lowering \(\delta\) results in applying a smaller watermark logit to these chosen tokens. This combination effectively weakens the watermark, as it biases only a few tokens with relatively lower strength, thereby encouraging the selection of the token with the highest model logit. Applying a weaker watermark to tokens following ADJs and DETs promotes the selection of the next token with the highest model logit, which is most likely to be a noun. This approach thereby enhances both semantic coherence and syntactic consistency.
Similarly, our method  allocates increased values of \(\gamma\) and \(\delta\) following punctuation (PUNCT) tokens.  Given the minimal constraints on subsequent tokens after PUNCT, as detailed in Appendix~\ref{sec:example}, our model potentially exploits this flexibility to embed a stronger watermark. 
It does so by putting more words into the green list (via  assigning  higher \(\gamma\) values) and  enhancing the preference for these tokens through elevated \(\delta\) values. }

\subsection{Robustness Against Attacks on Watermarks}

Considering that watermarked text can be easily altered to remove the watermark, thereby making detection challenging, 
 we evaluate the robustness of our method against two prevalent attacks:  1) the Paraphrase Attack~\cite{krishna2023paraphrasing}, 
 where watermarked text  is rephrased by another LLM, aiming to weaken the watermark while preserving the original semantics; 2) the Copy-Paste Attack~\cite{kirchenbauer2023reliability}, which  inserts watermarked text into its corresponding human-generated prompts used for its creation; this attack weakens the detection of watermarks as it increases the number of red tokens in the sentence. 

\paragraph{Paraphrase Attack.}

We utilized the Dipper paraphrase model~\cite{krishna2023paraphrasing} to perform paraphrase attacks. The model is finetuned from the T5-XXL~\cite{raffel2020exploring} model of 11B parameters, which is larger than the OPT-1.3B model we used for text generation. The larger model size results in improved generation capabilities~\cite{brown2020language}, ultimately leading to a more potent attack. We set the paraphrase strength to the same level as recommended in the Dipper GitHub repository for watermark detection, specifically lex=40, div=100.

\begin{figure}[t]
    \centering
    \includegraphics[width=0.45\textwidth]{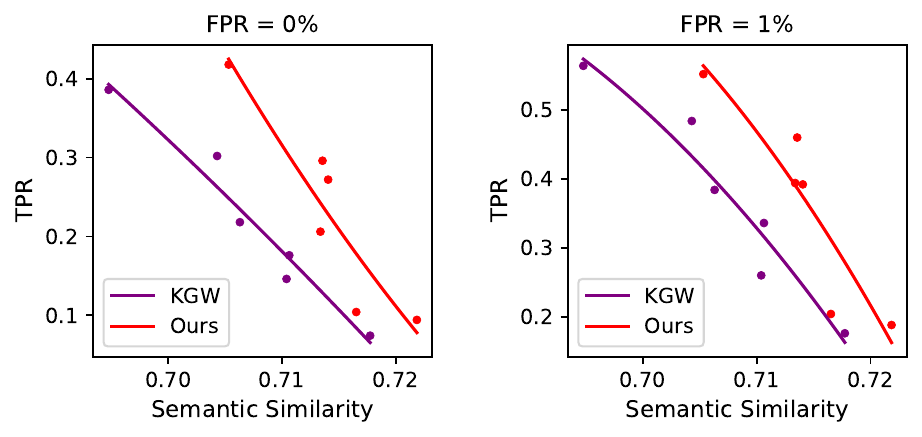}
    \caption{Comparison of our method with KGW under the Dipper paraphrase attack.}
    \label{fig:dipper}
    \vspace{-0.1cm}
\end{figure}

Figure~\ref{fig:dipper} shows that our method is more robust  against paraphrase attacks   than KGW, as evidenced by its superior Pareto frontier. 
This  can be attributed to the fact that paraphrasing tends to preserve local lexical structures, including punctuation patterns. As discussed in Sec.~\ref{analysis-of-delta-gamma},  our approach inserts  strong watermarks around punctuations. These watermarks remain intact even after paraphrasing, ensuring the detectability of the watermark in the altered text. Such a mechanism is lacking in KGW. 

\paragraph{Copy-Paste Attack.}
Our method is evaluated against two types of copy-paste attacks:  Copy-Paste-1 and Copy-Paste-3. In Copy-Paste-1, the entire watermarked text is inserted at a random position within the prompt. In Copy-Paste-3, the watermarked text is split into three segments, each of which is then randomly inserted at different positions in the prompt. The detection process is conducted in a similar fashion as in \citet{kirchenbauer2023reliability}, using a sliding window to compute the maximum z-score across text subsequences. The window size is 200 for Copy-Paste-1 and 60 for Copy-Paste-3. 

\begin{figure}
    \centering
    \begin{subfigure}[t]{0.45\textwidth}
        \includegraphics[width=\textwidth]{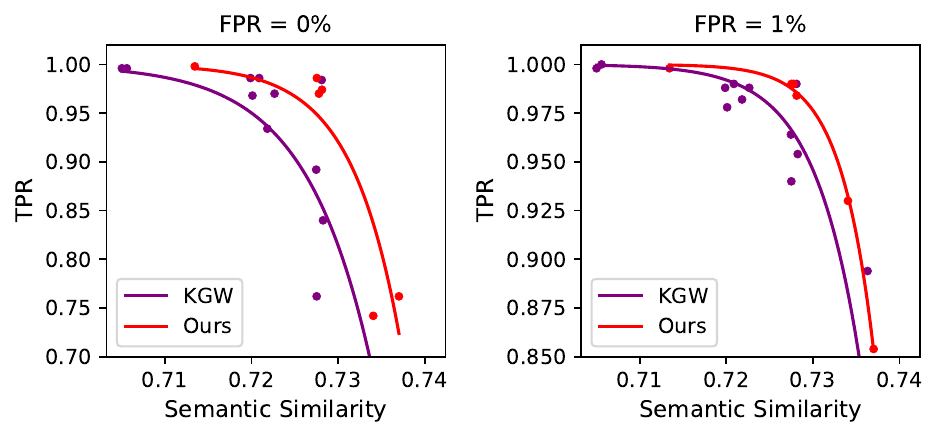}
        \label{fig:cp-1}
    \end{subfigure}
    \begin{subfigure}[t]{0.45\textwidth}
        \includegraphics[width=\textwidth]{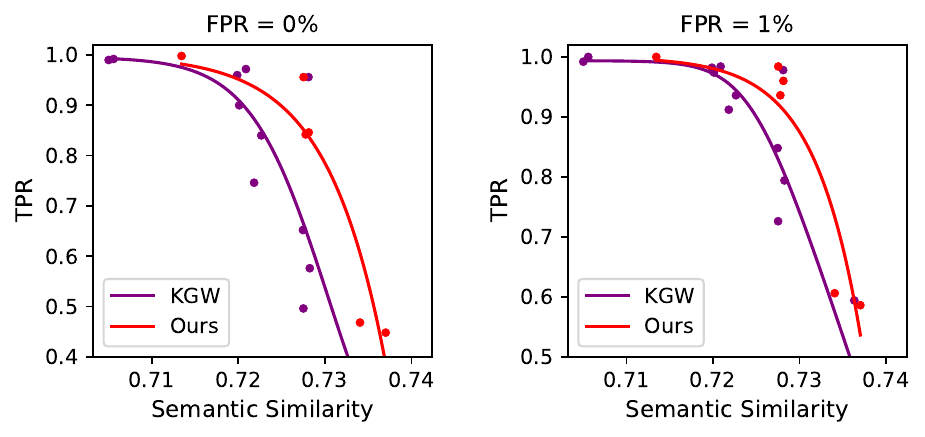}
        \label{fig:cp-3}
    \end{subfigure}
    \caption{Comparison of our method with KGW under the Copy-Paste-1 (the two figures at the top) and Copy-Paste-3 attack (the two figures at the bottom).}
    \label{fig:cp}
    \vspace{-0.2cm}
\end{figure}
 
As shown in Figures~\ref{fig:cp}, our method outperforms KGW in achieving a superior Pareto frontier across both scenarios. This improvement is largely due to the enhanced z-score, a direct result of our multi-objective optimization framework, which effectively  increases the number of green tokens in the watermarked text. This  ensures successful watermark detection even when human texts are inserted. 

\section{Conclusion}
In this work, we introduce a novel multi-objective optimization framework for watermarking LLMs during inference time. This method is designed to simultaneously optimize two light-weight networks, responsible for generating token-specific splitting ratios and watermark logits. The core objective of this approach is to minimize both detection loss and semantic loss, striving to find a Pareto optimal solution that enhances watermark detectability while preserving the semantic integrity of the generated text. Experiments demonstrate that our method consistently improves the Pareto frontier, surpassing previous techniques by offering improved watermark detectability and semantic integrity concurrently. Furthermore, our approach exhibits enhanced robustness against strong attacks, such as paraphrasing and copy-paste attacks, highlighting its practical effectiveness in safeguarding LLM outputs.

\section*{Impact Statement}
Our watermarking method for LLMs carries significant societal implications. By improving the detection of machine-generated texts, our approach has the potential to prevent the usage of LLMs in harmful activities, such as election manipulation campaigns, the dissemination of fake news, and academic dishonesty. Furthermore, when used responsibly, watermarking algorithms contribute to the protection of intellectual property rights, benefiting companies and content creators, and reducing the risk of unauthorized use. However, we realize the ethical considerations associated with our work, as the misapplication of these tools could mistakenly label human-generated content as LLM-generated, potentially leading to accusations of plagiarism against innocent individuals.

\bibliography{ref}
\bibliographystyle{icml2024}


\input{appdix}

\end{document}

%% file: appdix.tex
\newpage
\appendix
\onecolumn

\section{Post-Hoc Text Detection} 
\label{sec:aux-related-works}
Here is a supplement of related works, specifically focusing on the literature for post-hoc text detection.
Post-hoc text detection aims to distinguish generated texts from human-authored texts by analyzing generated texts without access to the LLMs (i.e., treating them as black boxes). These methods primarily leverage features extracted using external language models or by training models to act as detectors. \citet{gehrmann2019gltr} proposes the use of statistical metrics such as perplexity, entropy, and n-gram frequency to detect LLM-generated texts, as those metrics for generated texts may be different from human-written texts. Given a target text $S$, Sniffer~\cite{li2023origin} uses multiple accessible language models to compute a list of perplexity for the given text $S$, and train a linear classifier based on the perplexity features. and LLMDet~\cite{wu2023llmdet} rely on perplexity-based features for detection. In addition, there are supervised learning methods where classifiers are trained to specifically distinguish between texts generated by humans and LLMs. \citet{frohling2021feature} and \citet{solaiman2019release} utilized SVMs and regressions based on statistical features for LLM-generated text detection, while \citet{rodriguez2022cross} and \citet{zhong2020neural} employed neural networks for this purpose. Other works, such as those by \citet{solaiman2019release}, \citet{ippolito2019automatic}, \citet{guo2023close}, \citet{yu2023gpt}, and the OpenAI Detector~\cite{openai-detector}, involve fine-tuning a pre-trained RoBERTa~\cite{liu2019roberta} classifier for the detection of LLM-generated texts. 

However, the diverse and evolving nature of LLM-generated texts presents challenges in developing robust post-hoc detection techniques. For instance, the detection strategies effective for GPT-2 may not be applicable to GPT-3, highlighting the evolving complexity of these models~\cite{{gambini2022pushing}}. Additionally, these detection models are susceptible to adversarial attacks, which can deteriorate their performance \cite{wolff2020attacking}. A significant challenge in post-hoc text detection is the minimal difference between LLM-generated and human-authored texts, often leading to human content being mislabeled as LLM-generated \cite{liang2023gpt}. To mitigate these issues, watermarking approaches have been proposed that embed statistical signals during text generation that help distinguish it from human generated texts. These methods reduce false positive rates and substantially improve detection capabilities.

\section{Proof of Theorem~\ref{theorem:1}}

The proof of Theorem \ref{theorem:1} is detailed in \citet{cuzzocrea2021lyapunov}. The theorem is applicable under the condition that \( 0 < \mu_t < 1 \) $\forall t \in 1, \ldots, T$; that is, \( \mu_t \neq 0\) and \( \mu_t \neq 1\). This assumption is valid in our context since \( \gamma_t \) is neither 0 nor 1 for any \( t \). Specifically, \( \gamma_t = 0 \) would imply the absence of any tokens in the green list, indicating that watermarking has not been applied, whereas \( \gamma_t = 1 \) would suggest that every token is in the green list, rendering detection infeasible.

\section{Multiple-Gradient Descent Algorithm}\label{sec:appdix_moo}

As explained in Sec.~\ref{sec:moo}, we jointly optimize over the detection loss $L_D$ and semantic loss $L_S$ using multiple-gradient descent algorithm (MGDA)~\cite{desideri2012multiple, NeurIPS2018_Sener_Koltun}, which is formulated as: 
\begin{equation*}
    \min_{G_\gamma, G_\delta} L_D(G_\gamma, G_\delta)\text{ and }\min_{G_\gamma, G_\delta} L_S(G_\gamma, G_\delta).
\end{equation*}
In MGDA, the gradients of $L_D$ and $L_S$ with respect to $(G_\gamma, G_\delta)$ are firstly computed, which are denoted as $\boldsymbol{g}_D$ and $\boldsymbol{g}_S$, respectively. The resultant gradient vector, $\boldsymbol{g}$, is then estimated using $\boldsymbol{g}_D$ and $\boldsymbol{g}_S$ that directs the  optimization towards Pareto optimal solutions. This is determined as the minimum norm point within the convex hull formed by $\boldsymbol{g}_D$ and $\boldsymbol{g}_S$, which is formulated as follows:
\begin{equation}
\begin{array}{ll}
\lambda^* = \operatorname*{arg\,min}_{\lambda \in [0,1]} \|\lambda \boldsymbol{g}_D + (1 - \lambda) \boldsymbol{g}_S\|_2
\end{array}
\label{eq:gamma_moo}
\end{equation}

Note that the closed form of \(\lambda^*\) can be obtained using a few simple operations following~\citet{NeurIPS2018_Sener_Koltun}.
\[
\lambda^* = \begin{cases} 1, & \text{if } \boldsymbol{g}_D^T\boldsymbol{g}_S \geq \boldsymbol{g}_D^T\boldsymbol{g}_D \\ 0, & \text{if } \boldsymbol{g}_D^T\boldsymbol{g}_S \geq \boldsymbol{g}_S^T\boldsymbol{g}_S \\ \frac{(\boldsymbol{g}_S-\boldsymbol{g}_D)^T\boldsymbol{g}_S}{\|\boldsymbol{g}_D-\boldsymbol{g}_S\|^2}, & \text{otherwise} \end{cases}
\]
Using the obtained optimal $\lambda^*$, we estimate the resultant gradient direction as the convex combination of $\boldsymbol{g}_D$ and $\boldsymbol{g}_S$. 
\begin{equation}
\begin{array}{ll}
\boldsymbol{g} = \lambda^*\boldsymbol{g}_D + (1-\lambda^*)\boldsymbol{g}_S
\end{array}
\label{eq:grad_moo}
\end{equation}
Then at each step, the parameters of $(G_\gamma, G_\delta)$ are updated using the resultant gradient vector, $\boldsymbol{g}$, to optimize towards Pareto optimal solutions. This process is repeatedly executed until the end of total number of epochs. Theoretically, Multiple-Gradient Descent Algorithm (MGDA) is proven to converge to a Pareto stationary solution~\cite{desideri2012multiple, NeurIPS2018_Sener_Koltun}.

\section{Experimental Details}\label{sec:appdix_exp}

\paragraph{Training Details.}

A two-layer multilayer perceptron (MLP) is used as the \(\gamma\) and \(\delta\) generators, with the hidden layer dimension set to 64 and LeakyReLU as the hidden layer activation function~\cite{maas2013rectifier}. To ensure output values are within the 0 to 1 range, the final layer of the \(\gamma\)-generator incorporates a Sigmoid function. The weights of the MLPs are the only parameters we update during training. These weights were initialized using the Kaiming method~\cite{he2015delving}, and the Adam optimizer~\cite{kingma2014adam} was utilized for optimization. The training process involved a batch size of 8 and spanned 2 epochs with a fixed learning rate $1e-4$.  We set the temperature in Gumbel Softmax to be $0.1$. Every 100 steps, a checkpoint was saved, and the best-performing checkpoint on the validation set - judged in terms of improved detection and semantic coherence - was selected for the final evaluation. The reported results reflect the performance of this selected checkpoint on the test set, and we perform the inference using a batch size of 20.

\paragraph{Evaluation Details on Llama-2.} For Llama-2 7B and 13B, we directly loaded the original model from Hugging Face library. For Llama-2 70B, we loaded a 4-bit quantized model~\footnote{\url{https://huggingface.co/TheBloke/Llama-2-70B-GPTQ}} to fit the model to a single GPU.

\paragraph{Hardware.} The experiments are performed on Nvidia A100 GPUs with 80 GB of memory, including training and inference. Each experiment is run on a single GPU without model or data parallelism. We load the LLMs directly from huggingface~\cite{wolf2019huggingface} in \texttt{float16}, and use \texttt{PyTorch autocast} to train our MLPs in full precision. Each training instance takes 20 hours using 60 GB of memory. 

\paragraph{Curve-Fit.}
For all our results, we plot the curves given the points using \texttt{curve\_fit} under Python \texttt{scipy} module.
We use the five-parameter logistic curve~\cite{gottschalk2005five} and the exponential curve to fit the points.
\begin{equation}
    y=d + \frac{a-d}{\left(1+(x/c)^b\right)^g}, \ \ y=-a*e^{bx}+c.
\end{equation}
Here $(x, y)$ are the points we trying to fit, and the letters $a,b,c,d,g$ are parameters to fit the points. Initially, we apply the first function, which has a strong expressive ability, allowing it to fit a wide range of curve shapes. However, due to this high expressiveness and the limited number of data points, it may not always yield a concave curve. If so, we then resort to the second function, which is less versatile compared to the first, but its constrained fitting capability makes it more suitable for ensuring a concave curve shape in cases where data points are limited.

\paragraph{Trade-off Curves Between TPR and Semantic Similarity.} 
The trade-off curve between true positive rate (TPR) and semantic similarity at a specified false positive rate (FPR) is plotted for our method and the baselines for comparison. In this section, we outline the procedure for plotting the trade-off curves for different methods studied in this work. For each method, we plot this curve by varying the parameters that directly influence TPR and semantic similarity among their \textcolor{black}{appropriate} choices.

\begin{figure}[t]
    \centering
    \includegraphics[width=0.6\textwidth]{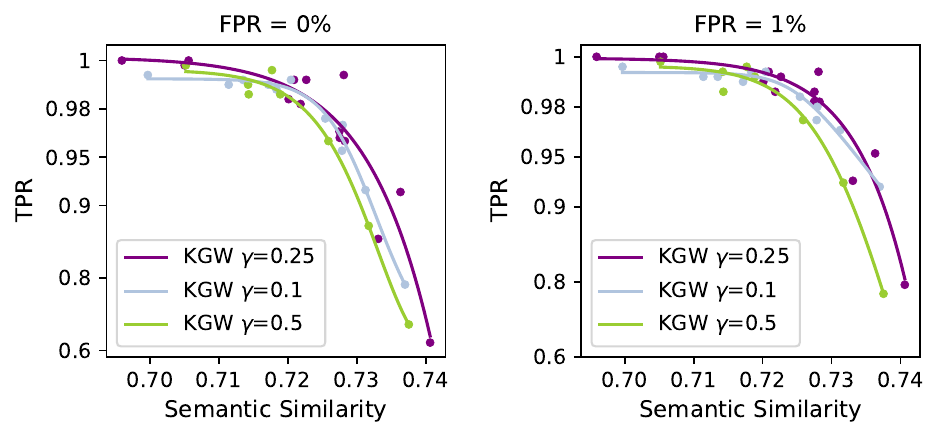}
    \caption{Fix $\gamma=\{0.1, 0.25, 0.5\}$ and varying $\delta$ for KGW.}
    \label{fig:kgw_init}
\end{figure}

For KGW, we identified \(\gamma=0.25\) as optimal and varied \(\delta\) in \([1.0, 3.0]\) to ensure a strong watermark, i.e., TPR \(> 0.6\) at a low FPR. We concluded that \(\gamma = 0.25\) was the best choice among the possible values of \(\{0.1, 0.25, 0.5\}\) specified in the paper, based on the following analysis: We fixed \(\gamma\) at one of the values in \(\{0.1, 0.25, 0.5\}\), varied \(\delta\) in the range of \([1.0, 3.0]\), and plotted their corresponding TPR and semantic similarity values, as shown in Figure~\ref{fig:kgw_init}. We observed that the curve corresponding to \(\gamma = 0.25\) was slightly higher than the others. Therefore, we concluded that \(\gamma = 0.25\) is relatively better, a finding also mentioned in their GitHub repository. Consequently, we identified the combination of \(\gamma = 0.25\) and \(\delta\) in the range of \([1.0, 3.0]\) as Pareto optimal. We then varied these parameters to obtain their corresponding TPR and semantic similarity values for KGW on the test set.

\textcolor{black}{For the SWEET and $\text{SWEET}_{\text{NoPrompt}}$ methods, we set \(\gamma\) at 0.25 and varied \(\delta\) within the range of \([1.0, 8.0]\). We explored higher \(\delta\) values to improve detectability by favoring green token selection. However, even with increased \(\delta\), a 100\% TPR at 0\% FPR was not achieved. Furthermore, we did not adopt their entropy threshold value directly, as it is specifically tailored to code generation. Instead, we calculated a new threshold tailored to our training dataset, using the same method they described. Specifically, for each token in a training example, we measured the entropy of generating the next token using LLM. These measurements were then used to estimate the entropy distribution. Finally, we set the threshold as the mean of this distribution. By varying these $\delta$ parameters, we obtained corresponding TPR and semantic similarity values on the test set, which were then plotted to create the curve for SWEET. Similarly, for SIR and $\text{SIR}_{\text{NoPrompt}}$, we varied \(\delta\) within the range of \([0.6, 2.0]\). They are more sensitive to the $\delta$ parameter. For SIR,  during watermark detection, we use prompts and the generated completions to get watermark logits, and calculate z values by averaging the watermark logits of the generated tokens. For $\text{SIR}_{\text{NoPrompt}}$, we only use the generated completions to get watermark logits and calculate z values.}

MultiBit is designed to encode a multi-bit message into generated text, with successful watermark detection occurring when the decoded message matches the embedded watermark. The message length is strategically chosen to maintain a predetermined FPR. We determined the FPR by calculating the percentage of human-written texts that were incorrectly decoded as containing the watermark message. We utilized the smallest message lengths that yielded FPRs no higher than 0\% and 1\%, specifically 9 and 7 bits, respectively. This baseline does not employ a splitting ratio $\gamma$. For the watermark logit $\delta$, we vary in the range $[0.9, 10.0]$.

EXP-edit uses exponential minimum sampling which is an approach to sample from a multinomial distribution. We set the temperature to the default value of 1. Additionally, our method and other baselines employ a Top-$k$ value of 50 during generation. However, the original EXP-edit generation process did not include the Top-$k$ feature, so we modified EXP-edit to incorporate this feature for a fair comparison. We refer to this modified version as EXP-edit (Top-$k$=50). It is important to note that indistinguishable methods like EXP-edit, which are based on pseudo-random sampling during generation, cannot easily extend to other decoding strategies like greedy sampling or beam search, where no randomness is involved.

In our method, we initialize the parameters $\gamma_t$ and $\delta_t$ to the values in $\{(0.1, 1.0)$, $(0.25, 1.0)$, $(0.25, 1.25)$, $(0.25, 1.5)$, $(0.25, 1.75)$, $(0.25, 2.0)\}$ for all $t$. These initializations represent the six best pairs of $\gamma$ and $\delta$ as shown in Figure~\ref{fig:kgw_init}. We further optimize them within our multi-objective framework to improve detectability and semantic coherence. The TPR and semantic similarity corresponding to these pairs are plotted on the test set to generate the performance curve for our method.

\section{Results on PPL and Z-Score}\label{appdx:ppl}

\begin{figure}[t]
    \centering
    \includegraphics[width=0.6\textwidth]{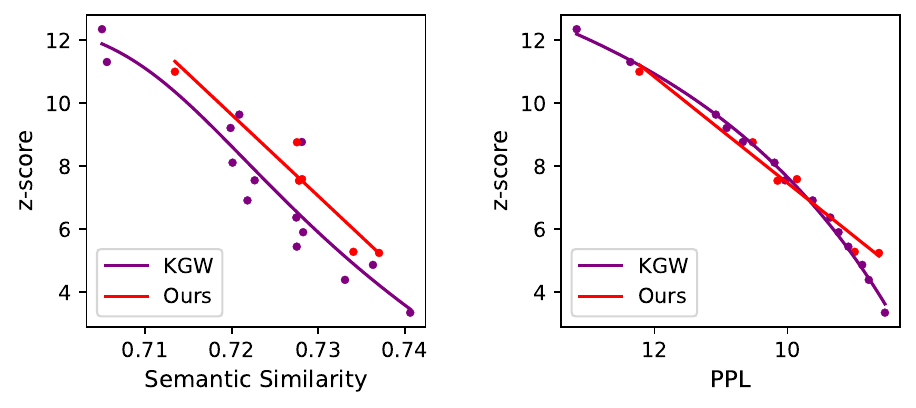}
    \caption{The trade-off between average z-score and semantic similarity (left) and oracle model perplexity (right).}
    \label{fig:z-score}
\end{figure}

Figure~\ref{fig:z-score} illustrates the trade-off between the average z-score and SimCSE performance, as well as between the average z-score and the oracle model perplexity (PPL), measured by OPT-2.7B. In Figure~\ref{fig:main_result}, we have already demonstrated that our model achieves a superior Pareto frontier in terms of TPR and SimCSE. Extending this finding, Figure~\ref{fig:z-score} shows that our model also surpasses KGW in both average z-score and SimCSE metrics. Additionally, in terms of PPL, our method performs comparably to the baseline methods.

\section{Results Based on Theoretical FPR}

\begin{figure}[t]
    \centering
    \includegraphics[width=0.6\textwidth]{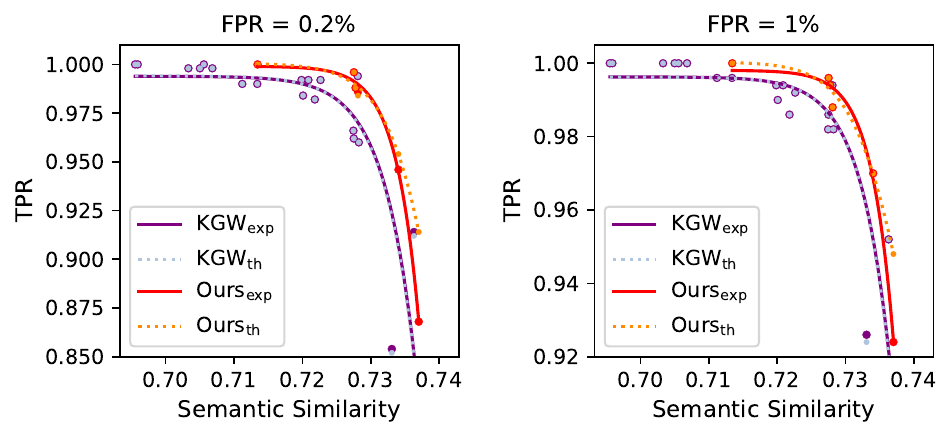}
    \caption{\textcolor{black}{
    Comparison of our method with KGW using one-sided z-score thresholds estimated with theoretical FPR (denoted by the subscript `th'). The results under empirical FPR (denoted by the subscript `exp'), shown in Figure~\ref{fig:main_result} of the main paper, are also included for comparison. The empirical and theoretical FPR curves are closely aligned or overlapping in most regions, indicating that they are close estimates.}}
    \label{fig:KGW_theoretical_FPR}
\end{figure}

We computed the theoretical FPR of our method based on the assumption used in KGW that the length of the generated text is sufficiently large. Under this assumption, the Poisson binomial distribution in our method can be approximated by a normal distribution according to the central limit theorem, enabling the use of a z-test to determine the theoretical FPR. Similarly, KGW calculates the theoretical FPR by approximating its binomial distribution with a normal distribution. Given FPR of 0.2\% and 1\%, the corresponding one-sided z-scores under the normal distribution are 2.88 and 2.33, respectively. The experimental results corresponding to these theoretical FPR are shown in Figure~\ref{fig:KGW_theoretical_FPR}, where our approach demonstrated superior performance compared to KGW. It is important to note that the theoretical FPR calculations for both KGW and our method may not be entirely precise due to the approximations of distributions~\cite{fernandez2023three}.

Alongside the curves corresponding to theoretical FPR, we also present curves corresponding to the empirical FPR, as shown in Figure~\ref{fig:main_result} of the main paper. The empirical FPR are computed using 500 human-written texts. For KGW, we fixed \(\gamma=0.25\) as the splitting ratio, and the empirical thresholds for KGW were 2.82 and 2.17 for FPR of 0.2\% and 1\%, respectively. 
The empirical thresholds of our method, for the six different initializations \((0.25, 2.0), (0.25, 1.75), (0.25, 1.5), (0.25, 1.25), (0.25, 1.0), (0.1, 1.0)\), are as follows. At an FPR of 0.2\%, the thresholds are \{2.68, 2.74, 3.11, 3.14, 2.87, 3.30\}. At an FPR of 1\%, the thresholds are \{2.22, 2.24, 2.42, 2.31, 2.35, 2.70\}.
Each point in our method has its own empirical FPR because we learn different splitting ratios, which impact the detection results. We observe that both the empirical and theoretical FPR curves are very close or overlapping in most regions, indicating that they are close estimates.

\section{Further Discussion on SWEET}\label{sec:sweet-dis}
At \((\gamma, \delta) = (0.25, 3.0)\), an analysis of LLM-generated texts that SWEET fails to detect at 0\% FPR indicates that, on average, only 7 out of 200 tokens are high-entropy and suitable for SWEET watermarking. This limited number of watermarkable tokens diminishes SWEET's detectability, even at high \(\delta\). To illustrate this limitation, consider a hypothetical scenario: even if, in the best case scenario (which is rare), all 7 out of 7 tokens are identified as green tokens, the z-score calculated using the formula \((|s|_G - \gamma T) / \sqrt{T\gamma(1-\gamma)}\) is 4.58. As per SWEET's implementation, $T$ is the total number of high-entropy tokens, and $|s|_G$ represents the identified green list tokens among them. However, with a higher number of watermarkable tokens, say 70, even if 50 were identified as green tokens, the z-score would increase to 8.97. Thus, the limited number of high-entropy tokens in SWEET hampers its detectability, highlighting the need for more watermarkable tokens for effective watermarking.

\section{Further Discussion on EXP-edit}
\label{sec:exp-editvsours}

In this section, we delve into a detailed comparison of our method with EXP-edit. EXP-edit's generation process, which employs exponential minimum sampling, is pseudo-random. This pseudo-random sampling process enables sampling from a multinomial distribution. By using a watermark key, this pseudo-random process becomes deterministic. Detection of the watermark relies on an edit score~\cite{kuditipudi2023robust}, which measures the likelihood of the text being watermarked given this watermark key.

EXP-edit is considered an indistinguishable method because it does not bias the output LLM distribution towards specific tokens, instead exploiting the randomness of the sampling strategy for embedding the watermark. We examine this indistinguishable property using the perplexity (PPL) of the generated texts. We use an oracle model, OPT-2.7B, to compute PPL of generated text.
PPL can estimate whether the output text is sampled from the same probability distribution, as such samples will have closer PPL values. However, it is important to note that PPL is not a reliable indicator of semantic quality. For example, the authors of \citet{piet2023mark} found that PPL favored repeated texts, with one model producing a partial response followed by repeated `$l$' characters, which was preferred due to its lower PPL.

\begin{table}[t]
\centering
\caption{\textcolor{black}{Comparison of TPR and PPL for No Watermark, EXP-edit, and Our Method on OPT-1.3B. The temperature is set to 1. Top-$k$=0 indicates sampling from the entire vocabulary. 
Exponential minimum sampling is an approach to sample from a multinomial distribution.} 
}
\begin{tabular}{l l c c c}
\toprule
 \textbf{Method} & \textbf{Decoding Strategy} & \textbf{TPR @ 0\%} & \textbf{TPR @ 1\%} & \textbf{PPL} \\ 
\midrule
 \multirow{4}{*}{No Watermark} & Beam search (Num Beams=2) & - & - & 1.628\\ 
 & Greedy decoding & - & - & 1.761\\ 
 & Multinomial sampling (Top-$k$=50) & - & - & 8.210\\ 
 & Multinomial sampling (Top-$k$=0) & - & - & 13.241\\ 
 \midrule
 \multirow{2}{*}{EXP-edit} & Exponential minimum sampling (Top-$k$=50) & 0.968 & 0.996 & 9.602 \\ 
 & Exponential minimum sampling (Top-$k$=0) & 0.922 & 0.996 & 16.235 \\ 
 \midrule
 \multirow{3}{*}{Ours} & Beam search (Num Beams=2) & 0.994 & 0.994 & \textbf{1.847} \\ 
 & Greedy decoding & 0.984 & 0.986 & 2.140 \\ 
 & Multinomial sampling (Top-$k$=50) & \textbf{1.000} & \textbf{1.000} & 12.227  \\ 
\bottomrule
\end{tabular}
\label{table:ppl_tpr_comparison}
\end{table}

The indistinguishable property of EXP-edit is evident when comparing the PPL values. Under multinomial sampling from the entire vocabulary, No watermark generation (Top-$k$=0) has a PPL of 13.241, while EXP-edit has a PPL of 16.235, as shown in Table~\ref{table:ppl_tpr_comparison}. Similarly, with Top-$k$=50, the PPL of EXP-edit (9.602) is close to that of No watermark (8.210).

On the other hand, control over the decoding strategy is also necessary to meet specific goals. For instance, if the goal is to attain the lowest PPL, beam search or greedy decoding can be used. As shown, No watermark with beam search, with number of beams 2, achieves the lowest PPL of 1.628, followed by greedy decoding with a PPL of 1.761. Our method, based on KGW, biases the distribution towards a specific set of tokens for embedding the watermark and can be applied on top of any decoding strategy as a heuristic. As demonstrated, our method with beam search decoding (num beams = 2) achieves a PPL of 1.847, and with greedy decoding, a PPL of 2.140, outperforming both variants of EXP-edit. Additionally, our method achieves good TPR values at both 0\% and 1\% FPR. In contrast, it is not straightforward to extend EXP-edit to greedy and beam search, which do not involve randomness.

Furthermore, the authors of \citet{piet2023mark} claim that indistinguishable methods are overly restrictive, and KGW-based methods offer more freedom to the user, enabling better detectability without significant loss of text quality.

\section{Further Analysis on Influence of Preceding Tokens on Watermark Strength}
\label{sec:example}
In Sec.~\ref{analysis-of-delta-gamma}, we analyze the learned watermark logits, $\delta$, and splitting ratios, $\gamma$. Our analysis indicates that lower \(\gamma\) and \(\delta\) values (weaker watermark) are assigned to adjectives (ADJ) and determiners (DET), likely due to the high likelihood of a noun following. In contrast, punctuation (PUNCT) is assigned higher \(\gamma\) and \(\delta\) values (stronger watermark) possibly due to the absence of restrictions on the subsequent token. To further support this claim, we estimated the transition probabilities from adjectives, determiners, and punctuation to the next token's part of speech (POS) tag using our training dataset. Notably, our observations show that a determiner is followed by a noun with a 0.7 probability, while an adjective is followed by a noun with a 0.56 probability. In the case of punctuation, we observed a nearly uniform distribution over the next token's POS tags. These observations reinforce our claims in Sec.~\ref{analysis-of-delta-gamma}.

\section{Qualitative Analysis}
\input{last_example}

We present a qualitative analysis of an example randomly selected from the test set in Table~\ref{tab:example}. Using the same prompt, we compare the no-watermark generation, KGW generation, and our model's generation. We observe that our method has a higher z-score (6.45) compared to the one obtained from the KGW method (5.77). Additionally, the SimCSE score is also higher for our method (0.85) compared to KGW (0.73).

We can also qualitatively observe the impact of the token-specific watermarking employed by our method. For example, consider the sequence `...would apply to all commercial...' from the third line of our method. Here, `apply' follows `would,' `to' follows `apply,' and `all' follows `to,' with each of these tokens being marked as green. In contrast, the token succeeding `all,' i.e., `commercial,' is marked as red. Upon calculating the splitting ratio (\(\gamma\)) and the watermark logit (\(\delta\)) for `would,' `apply,' `to,' and `all,' we find that `all' is the only token associated with both a low \(\gamma\) and a low \(\delta\). This demonstrates that our method adapts its watermarking strength to maintain semantic coherence for significant words like `commercial,' which is present in the prompt. Conversely, for tokens like `apply,' `to,' `all,' our model enforces a stronger watermark, aiming to boost detectability.

On the other hand, the KGW method exhibits limitations in handling token specificity. For instance, examining the sequence `{...a single income property...}' from the second line of a KGW-generated text reveals this issue. KGW applies uniform \(\gamma\) and \(\delta\) values across all tokens, leading to the generation of a green token `{income},' which is less relevant to the main idea.

\begin{figure}[t]
    \centering
    \includegraphics[width=0.6\textwidth]{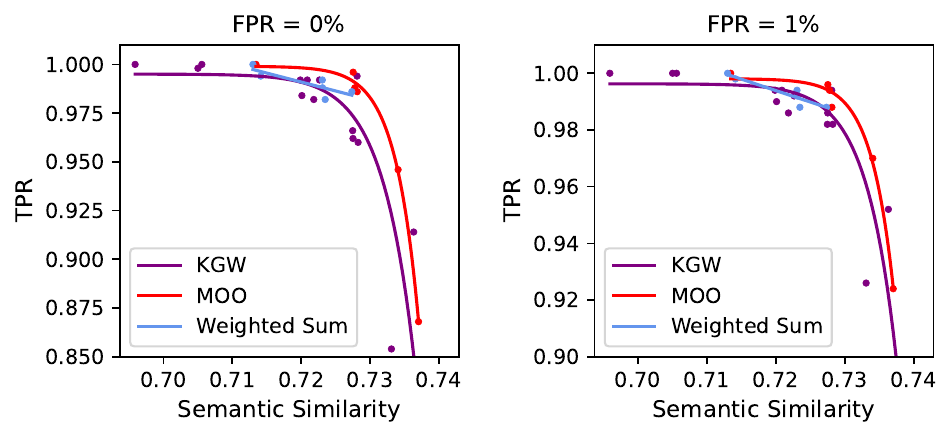}
    \caption{\textcolor{black}{Comparison of multi-objective optimization and weighted sum optimization. For the weighted sum optimization, we set \(\lambda_{ws}\) to \(4 \times 10^{-4}\) based on the averaged value of \(\frac{\lambda_{moo}}{1 - \lambda_{moo}}\) during the optimization.}}
    \label{fig:weighted}
\end{figure}

\begin{figure}[t]
    \centering
    \begin{subfigure}[t]{0.48\textwidth}
        \includegraphics[width=\textwidth]{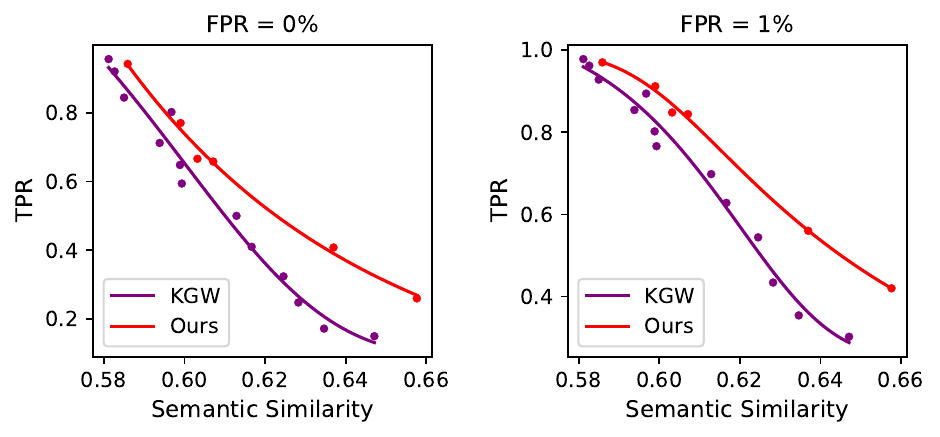}
        \caption{Generation length = 50 tokens.}
    \end{subfigure}
    \begin{subfigure}[t]{0.48\textwidth}
        \includegraphics[width=\textwidth]{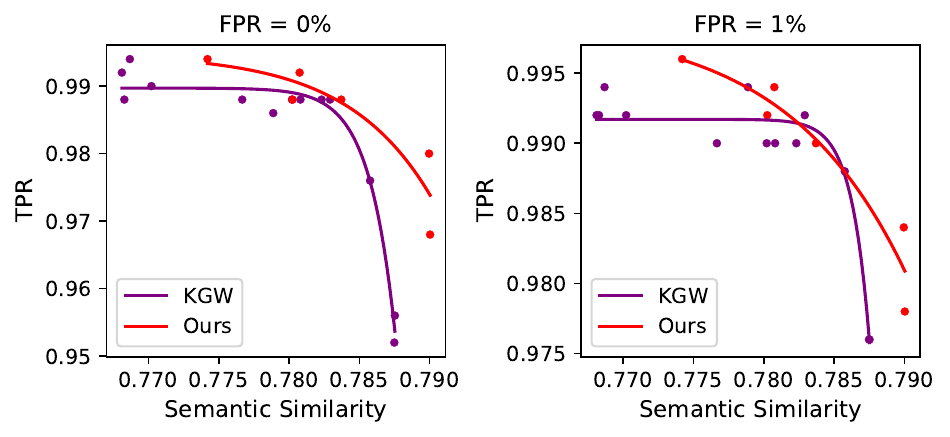}
        \caption{Generation length = 400 tokens.}
    \end{subfigure}
    \caption{\textcolor{black}{Comparison of our method with KGW on short and long text generations. For short text generations, we set the generation length to 50 tokens. For long text generations, we set the generation length to 400 tokens.}}
    \label{fig:opt_diff_length}
    \vspace{-0.2cm}
\end{figure}

\section{Ablation Study: Comparison of MOO and Weighted Sum Optimization}
\label{sec:appdix-weighted}

Multi-Objective Optimization (MOO) offers a more advanced approach than traditional weighted sum optimization for balancing various objectives. In contrast to weighted sum, which depends on a fixed hyperparameter to balance two objectives, MOO dynamically estimates the optimal gradient direction, effectively managing the trade-off between the objectives. 

To demonstrate the effectiveness of MOO, we compare it with the weighted sum approach. For weighted sum, our $\gamma$- and $\delta$-generators are trained on a single objective function that combines \( L_D \) and \( L_S \) using a trade-off parameter $\lambda_{ws}$, formulated as:
\begin{equation} 
    L = L_S + \lambda_{ws} L_D.
\end{equation}

In MOO, the optimal \( \lambda_{moo} \) is calculated for each iteration to balance the gradients from both objectives. This dynamic adjustment of \( \lambda \) leads us towards the Pareto optimal solution, showcasing MOO's superior capability in handling multiple objectives.

In this analysis, we set the \(\lambda_{ws}\) to \(4 \times 10^{-4}\). This value is based on our MOO experiment for a fair comparison. Although \(\lambda_{moo}\) is dynamically adjusted throughout the training process, we calculate its average value across this process and use it to determine \(\lambda_{ws}\). However, since the resultant gradient in MOO is a convex combination of gradients from individual objectives, using \(\lambda_{moo}\) as the coefficient, we set \(\lambda_{ws}\) as \({\lambda_{moo}}/{(1-\lambda_{moo})}\), which turns out to be the aforementioned value. 
We train with the following initialization:
$(0.25, 0.75)$, $(0.25, 1.0)$, $(0.25, 1.25)$, $(0.25, 1.5)$, $(0.25, 1.75)$, and $(0.25, 2.0)$.

The results in Figure~\ref{fig:weighted} show that 5 out of 6 points obtained by weighted sum optimization fall below the Pareto frontier established by MOO, demonstrating the effectiveness of MOO. These findings suggest that, for weighted sum optimization, improving performance for each initialization requires estimating the optimal \(\lambda_{ws}\) through trial and error, a process that can be exhaustive. However, by employing MOO, our method dynamically estimates this trade-off parameter, ensuring convergence towards the Pareto optimal solution.

\section{Computational Costs}
\begin{table}[t]
\small
\centering
\caption{\textcolor{black}{Memory utilization of OPT-1.3B for generating 200 tokens, measured in MB.}}
\begin{tabular}{l|c|c}
\toprule
\textbf{Method} & \textbf{Generation (MB)} & \textbf{Detection (MB)} \\
\midrule
No Watermark & 3475 & - \\
KGW & 3477 & 503 \\
SWEET & 3477 & 4933 \\
EXP-edit & 3486 & 416 \\
SIR & 5475 & 2425 \\
MultiBit & 4517 & 1485 \\
Ours & 3731 & 643 \\
\bottomrule
\end{tabular}
\label{tab:memory}
\end{table}

We evaluate the computational time of our method for both text generation and watermark detection, and compare it to the baselines. Table~\ref{tab:speed} shows the results. We also show memory utilization in Table~\ref{tab:memory}. Our method exhibits higher generation and detection speeds compared to EXP-edit, SIR, and MultiBit. The generation and detection time of our method is comparable to that of KGW, SWEET, and No Watermarking. Overall, our method achieves superior detectability and semantic coherence without a significant increase in computational cost or memory cost.

\section{Performance Across Varied Generation Lengths}\label{appdx:opt_diff_length}
In our main experiments, the generation length was set to 200 tokens. Here, we extend our evaluation to include both shorter and longer text generations. For short text generations, we configured the prompt length to at least 200 tokens, followed by a generation of 50 tokens by the LLM. For longer texts, the prompt was set to a minimum of 600 tokens, with the LLM generating 400 tokens. We used the C4 news-like dataset and the OPT-1.3B model for these text generations. As illustrated in Figure~\ref{fig:opt_diff_length}, our method is relatively better than KGW, showcasing the efficacy of our approach across various generation lengths.

\section{Performance on Different Datasets}
\begin{figure}
    \centering
    \begin{subfigure}[t]{0.48\textwidth}
        \includegraphics[width=\textwidth]{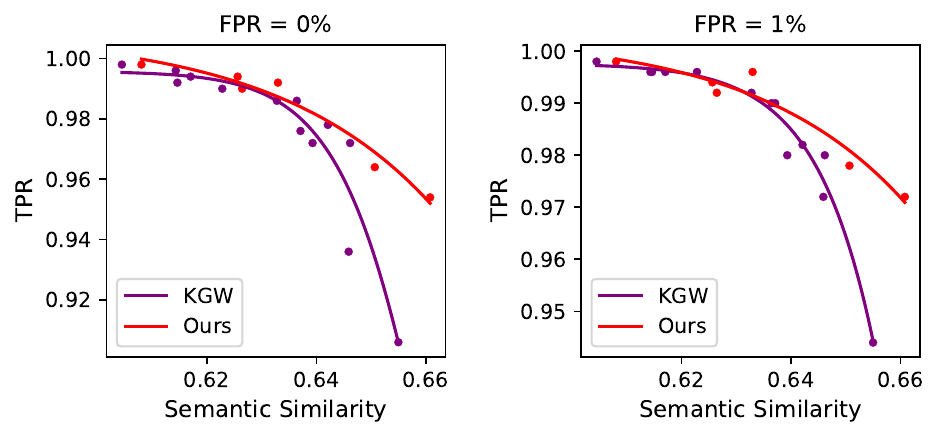}
        \caption{Essays dataset}
    \end{subfigure}
    \begin{subfigure}[t]{0.48\textwidth}
        \includegraphics[width=\textwidth]{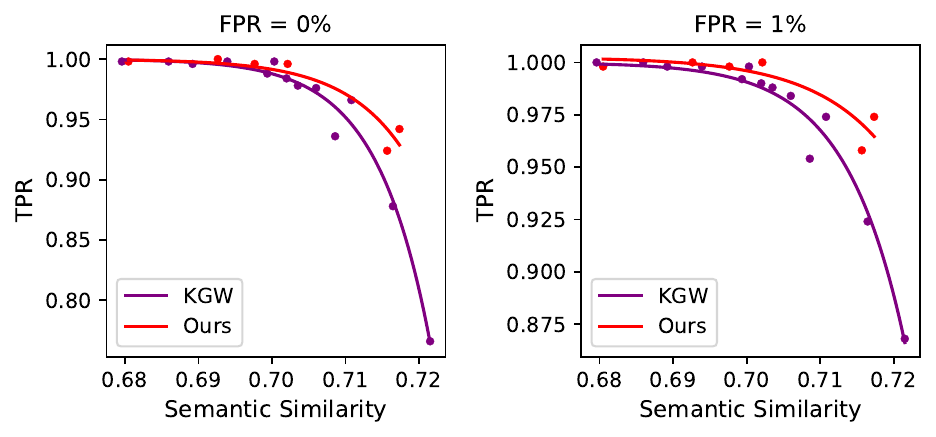}
        \caption{HC3 dataset}
    \end{subfigure}
    \caption{\textcolor{black}{Comparison of our method with KGW on the Essays and HC3 datasets.}}
    \label{fig:new_datasets}
\end{figure}

We further evaluated our learned models (\(\gamma\)- and \(\delta\)-generator networks), initially trained on the C4 dataset, on two additional datasets without further training. The Essays dataset~\cite{essays-with-instructions} comprises sets of instructions paired with corresponding essays. For this dataset, we used the instructions as prompts to generate essay responses.
The HC3 dataset~\cite{guo-etal-2023-hc3} contains ChatGPT-generated text, and we use the initial 100 tokens of the ChatGPT-generated text as prompts to generate the completions. 
The OPT-1.3B LLM was used for the experiment, with the generation length fixed at 200 tokens. As shown in Figure~\ref{fig:new_datasets}, our method outperforms the baseline KGW method, demonstrating the robust transferability of the \(\gamma\) and \(\delta\) networks across different datasets.

\section{Modified Corruption Attack}\label{appdx:attack}
\begin{figure}[t]
    \centering
    \begin{subfigure}[t]{0.49\textwidth}
        \includegraphics[width=\textwidth]{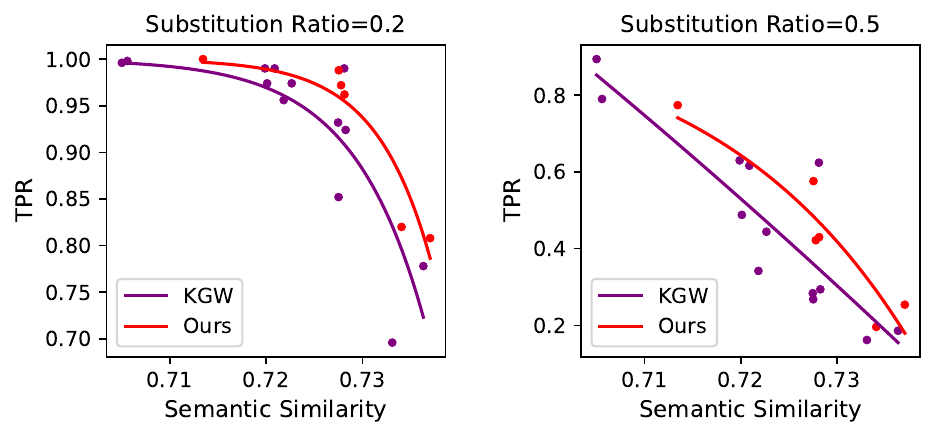}
        \caption{Substitution using low-$\gamma$ tokens.}
    \end{subfigure}
    \begin{subfigure}[t]{0.49\textwidth}
        \includegraphics[width=\textwidth]{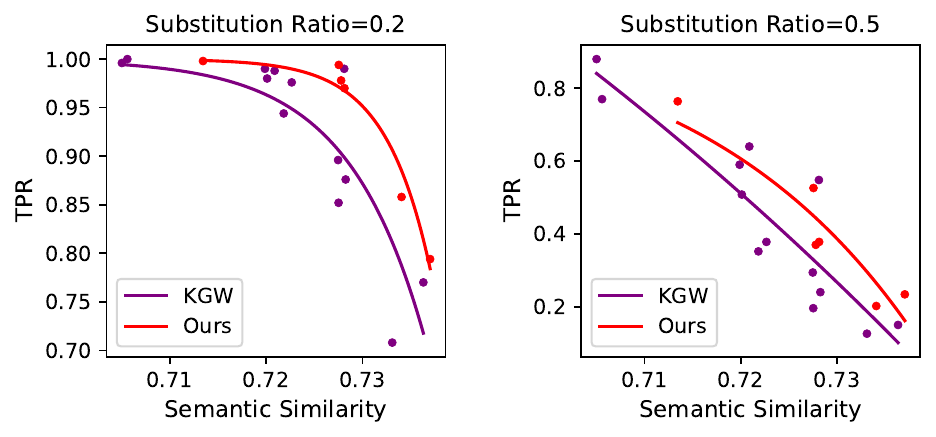}
        \caption{Substitution using high-$\gamma$ tokens.}
    \end{subfigure}
    \begin{subfigure}[t]{0.49\textwidth}
        \includegraphics[width=\textwidth]{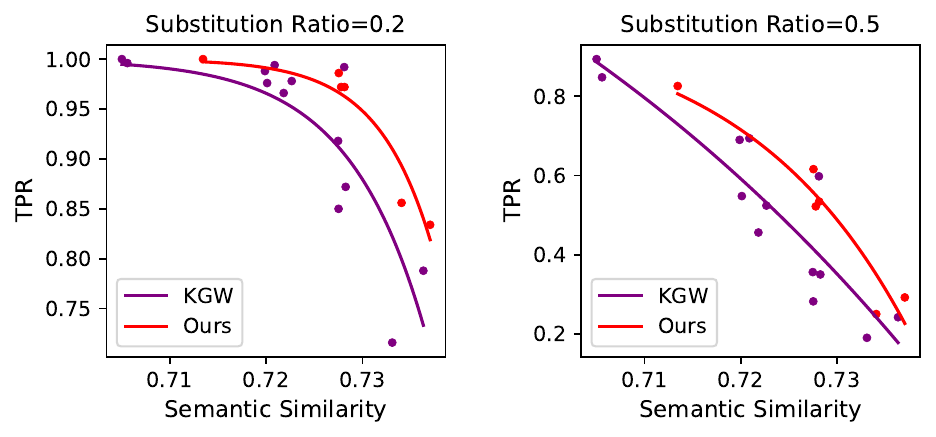}
        \caption{Substitution using random tokens.}
    \end{subfigure}
    \caption{\textcolor{black}{Corruption attack where the attacker substitutes P\% of tokens with low-$\gamma$, high-$\gamma$, and random tokens when the FPR is 1\%. P is set to 20\% and 50\%.}}
    \label{fig:gamma_attack}
\end{figure}

In this evaluation, we assume the attacker knows the \(\gamma\) value for each token as determined by our method, enabling them to perform a corruption attack based on this knowledge. To investigate how varying $\gamma$ values affect attack efficacy, we conducted experiments with three types of corruptions: 
\vspace{-0.5pc}
\begin{itemize}
    \itemsep0em 
    \item Using our trained $\gamma$ network to select 30 tokens in the vocabulary with the lowest $\gamma$ values; then randomly substituting P\% of the tokens in a watermarked text with tokens uniformly selected from this low-$\gamma$ list. 
    \item Using our trained $\gamma$ network to select 30 tokens in the vocabulary with the highest $\gamma$ values; then randomly substituting P\% of the tokens in a watermarked text with tokens uniformly selected from this high-$\gamma$ list.
    \item Randomly substituting P\% of the tokens in a watermarked text with tokens drawn uniformly from the vocabulary.
\end{itemize}
\vspace{-0.5pc}
We tested our method using substitution ratios (P\%) of 20\% and 50\%, and the results are shown in Figure~\ref{fig:gamma_attack}. Our method consistently outperformed KGW across all attack scenarios. This underscores the robustness of our method against Corruption Attack, even when the watermarked text is corrupted using tokens selected based on $\gamma$ values derived from our trained $\gamma$ network. The strong robustness of our method mainly stems from the improved z-score, achieved through our multi-objective optimization framework that concurrently maximizes the z-score and SimCSE. As a result, even when a Corruption Attack attempts to alter the $\gamma$ scores of certain tokens, the remaining tokens maintain a sufficiently high z-score, facilitating the easy detection of the watermark.

%% file: last_example.tex
\begin{table}[t]
    \centering
    \caption{Qualitative result on a randomly selected test example is presented. This includes the prompt, the no-watermark text, the watermarked text from the KGW method, and the watermarked text from our method. To assess detectability, we provide the z-scores for the no-watermark text, the KGW method, and our method. The z-score for the no-watermark text is calculated under the assumption that it was generated by our method. Additionally, we include SimCSE evaluations to measure the similarity between the text generated by the KGW and our methods, and the no-watermarked text.
    }
    \begin{tabularx}{\textwidth}{|Y|Y|}
    \toprule
        \textbf{Prompt} 
        
    $\cdots$
    Typically, when the assessed value goes up dramatically for a big multi-tenant commercial building, the property owner isn't the one on the hook for the bill — it’s the tenants.

    Sylvester, who previously sat on the B.C. Assessment Board and the Downtown Vancouver Business Improvement Association, said she’s been discussing the idea with various people recently, including, just in the last week, the former chair of B.C. Assessment, and a provincial cabinet minister.

    But property-tax agent Paul Sullivan said while the idea is “not without merit,” there are some
        & \textbf{No watermark (z-score: -0.40)} 

        \ issues\re{ with it, including who}\gr{ would}\re{ be responsible.}
        
        \re{``}\gr{Who is going}\re{ to be responsible for paying this? Who are the landlords}\gr{?}\re{'' he asked. ``Are}\gr{ they}\re{ just the large commercial property owners}\gr{?}\re{ Are they multi-ten}\gr{ant}\re{ buildings where the entire}\gr{ business}\re{ is owned by the landlord?''}
        
        \re{But}\gr{ Syls}\re{eter said the city could do it by creating}\gr{ a}\re{ new low-impact}\gr{ category.}\re{ Under her}\gr{ plan,}\re{ which she's}\gr{ presented}\re{ in a}\gr{ report}\re{, the existing triple-net category would become a}\gr{ low}\re{-impact}\gr{ category}\re{, which would}\gr{ allow small businesses to}\re{ still}\gr{ pay}\re{ the higher property rates.}
        
        \re{It}\gr{ will}\re{ provide}\gr{ relief to}\re{ the}\gr{ business community}\re{,}
        $\cdots$
        \\\midrule
         \textbf{KGW (z-score: 5.77, SimCSE: 0.73)} 

        \ questions\re{ about the size of}\gr{ the}\re{ proposed}\gr{ sub}\re{category.}
        
        \re{``To my understanding}\gr{ it}\re{'s going to be a single}\gr{ income property-}\re{income tax}\gr{ scheme,'}\re{' Sullivan said. ``}\gr{So as a result,}\re{ there are going to be}\gr{,}\re{ quite a few, many}\gr{ (business}\re{es) that}\gr{ probably would}\re{ fall to that part}\gr{ instead}\re{ of}\gr{ the}\re{ (mainstream)}\gr{ part.}\re{''}
        
        \re{For example,}\gr{ Sullivan noted,}\re{ under}\gr{ the}\re{ city's}\gr{ existing tax rate}\re{, a}\gr{ small-}\re{business}\gr{ operator would pay 80}\re{ per cent}\gr{ of the}\re{ property}\gr{-}\re{tax rate}\gr{ on an \$}\re{800,000 home, and}\gr{ 15 per}\re{ cent on}\gr{ anything}\re{ above}\gr{.}

        \re{But}\gr{ if}\re{ Sy}\gr{lves}\re{ter}\gr{ proposes a}\re{ new subcategory within}\gr{ that}\re{ tax bracket,}\gr{ that employee}\re{ would}\gr{ pay}\re{ less}\gr{ as a business has}\re{ a ``limited}\gr{ ability to grow}\re{ their}\gr{ cash}\re{.''} $\cdots$
        &
         \textbf{Ours (z-score: 6.45, SimCSE: 0.85)}
        
        \ questions\re{ about the size of}\gr{ the}\re{ proposed change}\gr{, what it}\re{ would mean to businesses that}\gr{ don}\re{'}\gr{t}\re{ have large}\gr{ landlords}\re{, and how}\gr{ it}\re{ would}\gr{ apply to all}\re{ commercial}\gr{-property}\re{ owners}\gr{.}
        
        \re{``}\gr{The biggest issue}\re{ is how}\gr{ it}\re{ would}\gr{ apply to}\re{ non}\gr{-residential}\re{ buildings}\gr{. If}\re{ you}\gr{ consider that the}\re{ vast}\gr{ majority}\re{ of Vancouver}\gr{ is}\re{ for residential}\gr{, what}\re{ are the implications of a change that would}\gr{ apply to}\re{ non}\gr{-residential}\re{ buildings only?''}\gr{ said Sullivan.}
        
        \re{But}\gr{ Sylves}\re{ter said}\gr{ her}\re{ proposal}\gr{ is}\re{ to put the property}\gr{-}\re{tax rate}\gr{ on small businesses}\re{ in the}\gr{ form}\re{ of an exemption, so}\gr{ businesses like}\re{ hers — which range from small bakeries}\gr{ to boutique}\re{ clothing}\gr{ shops — would not}\re{ pay triple-net}\gr{.}$\cdots$
         \\\bottomrule
    \end{tabularx}
    \label{tab:example}
\end{table}